\documentclass[acmtog,]{acmart}
\acmSubmissionID{85}

\usepackage{booktabs} 
\usepackage{xcolor}
\usepackage{float}
\usepackage{cuted}
\usepackage{capt-of}
\usepackage{amsmath}
\usepackage[inline]{enumitem}
\usepackage{subcaption}
\usepackage{multirow}
\usepackage[ruled]{algorithm2e} 

\SetAlFnt{\small}
\SetAlCapFnt{\small}
\SetAlCapNameFnt{\small}
\SetAlCapHSkip{0pt}

\acmJournal{TOG}

\setcopyright{acmlicensed}\acmJournal{TOG}
\acmYear{2021}\acmVolume{40}\acmNumber{4}\acmArticle{85}\acmMonth{8} \acmDOI{10.1145/3450626.3459792}
\begin{document}
\title{Capturing Detailed Deformations of Moving Human Bodies}

\author{He Chen}
\orcid{0000-0002-5819-3453}
\email{ankachan92@gmail.com}
\author{Hyojoon Park}
\orcid{0000-0002-3796-9777}
\email{hjoonpark.us@gmail.com}
\author{Kutay Macit}
\orcid{0000-0002-0591-1044}

\email{ucanbizon@gmail.com}
\author{Ladislav Kavan}
\orcid{0000-0001-8549-0878}
\email{ladislav.kavan@gmail.com }
\affiliation{
\institution{University of Utah}
 \streetaddress{Electrical and Computer Engineering
50 S. Central Campus Drive
MEB Lab 3335}
 \city{Salt Lake City}
 \state{UT}
 \postcode{84112}
 \country{USA}
}
\begin{abstract}
   We present a new method to capture detailed human motion, sampling more than 1000 unique points on the body. Our method outputs highly accurate 4D (spatio-temporal) point coordinates and, crucially, automatically assigns a unique \textit{label} to each of the points.
   The locations and unique labels of the points are inferred from individual 2D input images only, without relying on temporal tracking or any human body shape or skeletal kinematics models. 
   Therefore, our captured point trajectories contain all of the details from the input images, including motion due to breathing, muscle contractions and flesh deformation, and are well suited to be used as training data to fit advanced models of the human body and its motion.
   The key idea behind our system is a new type of motion capture suit which contains a special pattern with checkerboard-like corners and two-letter codes. The images from our multi-camera system are processed by a sequence of neural networks which are trained to localize the corners and recognize the codes, while being robust to suit stretching and self-occlusions of the body. Our system relies only on standard RGB or monochrome sensors and fully passive lighting and the passive suit, making our method easy to replicate, deploy and use. Our experiments demonstrate highly accurate captures of a wide variety of human poses, including challenging motions such as yoga, gymnastics, or rolling on the ground.
\end{abstract}

%
%

\ccsdesc[500]{Computing methodologies~Motion capture}

%
%
\keywords{human animation, motion capture, skin deformation}
\newcommand{\TODO}[1]{\textcolor{red}{ToDo:#1}}
\newcommand{\LK}[1]{\textcolor{orange}{LK: #1}}
\newcommand{\AC}[1]{\textcolor{blue}{AC: #1}}
\newcommand{\Rv}[1]{#1}
\newcommand{\ACChange}[1]{{#1}}
\newcommand{\JP}[1]{\textcolor{red}{JP: #1}}
\newcommand{\OUTLINE}[1]{ \textcolor{blue}{#1}}
\newcommand{\Fig}[1]{Fig. \ref{#1}}
\newcommand{\Cornerdet}{\textit{CornerdetNet}}
\newcommand{\Rejector}{\textit{RejectorNet}}
\newcommand{\RecogNet}{\textit{RecogNet}}
\newcommand{\AppendixDataAug}{Supplementary Material B.1}
\newcommand{\AppendixSynthData}{Supplementary Material B.2}

\begin{strip}
    \centering
    \setlength\belowcaptionskip{-3ex}
    \includegraphics[width=\textwidth]{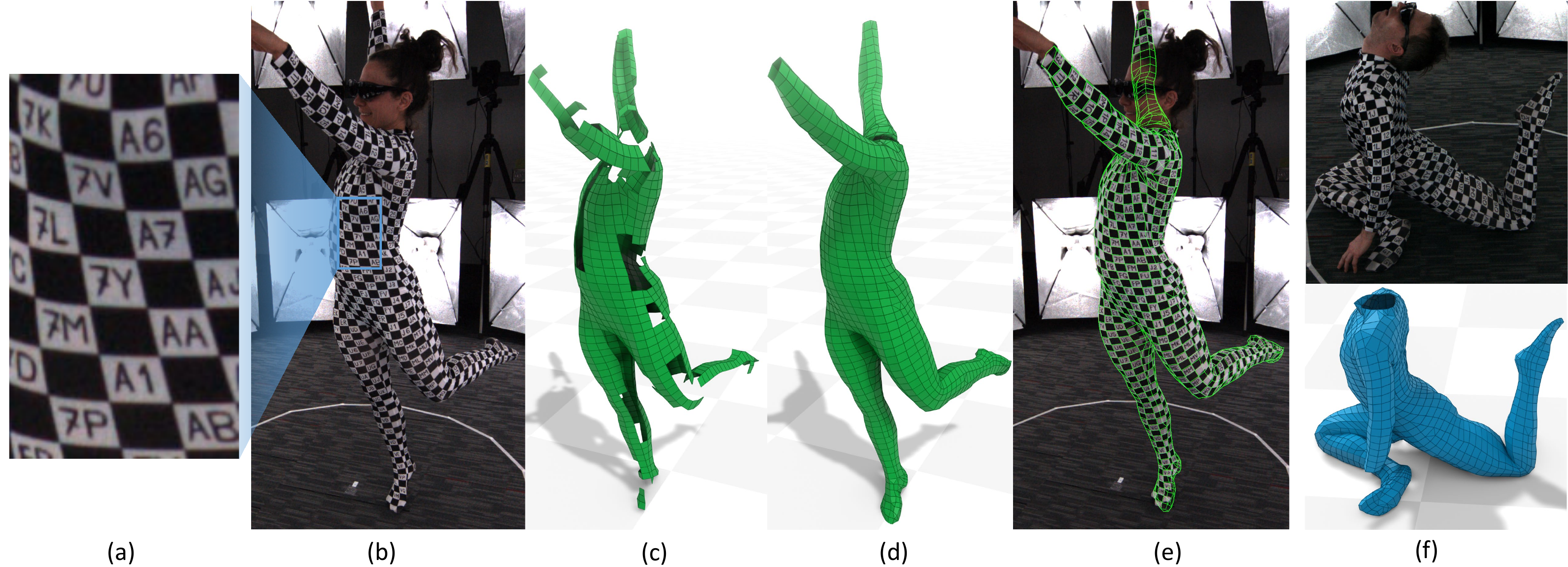}
    \captionof{figure}{ (a) Our novel motion capture suit with a special pattern; (b) An example input image from our multi-camera capture system; (c) Raw 3D reconstruction of \textbf{labeled} corners from the suit (raw data, no body model was used); (d) The result after interpolating missing observations (here, a body model is used); (e) Our reconstructed mesh aligns very closely with the original input images; (f) Our method works even in uncommon poses with many self-occlusions, such as this yoga pose.}
    \label{fig:teaster}
\end{strip}
\maketitle
\section{Introduction}

In most real-world images, the human body is occluded by clothing, making precise body measurements difficult or impossible.  A significant amount of previous work focuses on approximate but robust pose estimation in the wild \cite{cao2018openpose, guler2018densepose}.
However, a small muscle twitch or the speed of breathing may contain signals that are critical in certain contexts, e.g., in the context of social interactions, minute body shape motion may reveal important information about the person's emotional state or intent \cite{joo2018total}. Detailed human body measurements are highly relevant also in orthopedics and rehabilitation \cite{zhou2008human}, virtual cloth try on \cite{giovanni2012virtual, ma20autoenclother}, or building realistic avatars for telepresence and AR/VR \cite{barmpoutis2013tensor, lombardi2018deep}. 

When precise measurements are needed, prior work utilized either 1) reflective markers attached to a motion capture suit or glued to the skin \cite{park2006capturing}, or 2)  painting colored patterns on the skin \cite{bogo2017dynamic}.
The traditional reflective (``mocap'') markers present certain limitations. Because all of the markers look alike (Fig. \ref{fig:SuitDesign}a), marker labeling relies strongly on temporal tracking and high frame-rate cameras. However, robust marker labeling is a hard problem \cite{song2016fast} which often requires manual corrections, especially for markers that have been occluded for too long. \ACChange{The difficulty} of this problem grows with the number of markers \cite{park2006capturing}, thus sparse marker sets are most common in the industry. Sparse marker sets are sufficient for fitting a low-dimensional skeletal body model, but not for capturing the details of flesh deformation or motion due to breathing.

\begin{figure*}
    \centering
    \includegraphics[width=0.95\textwidth]{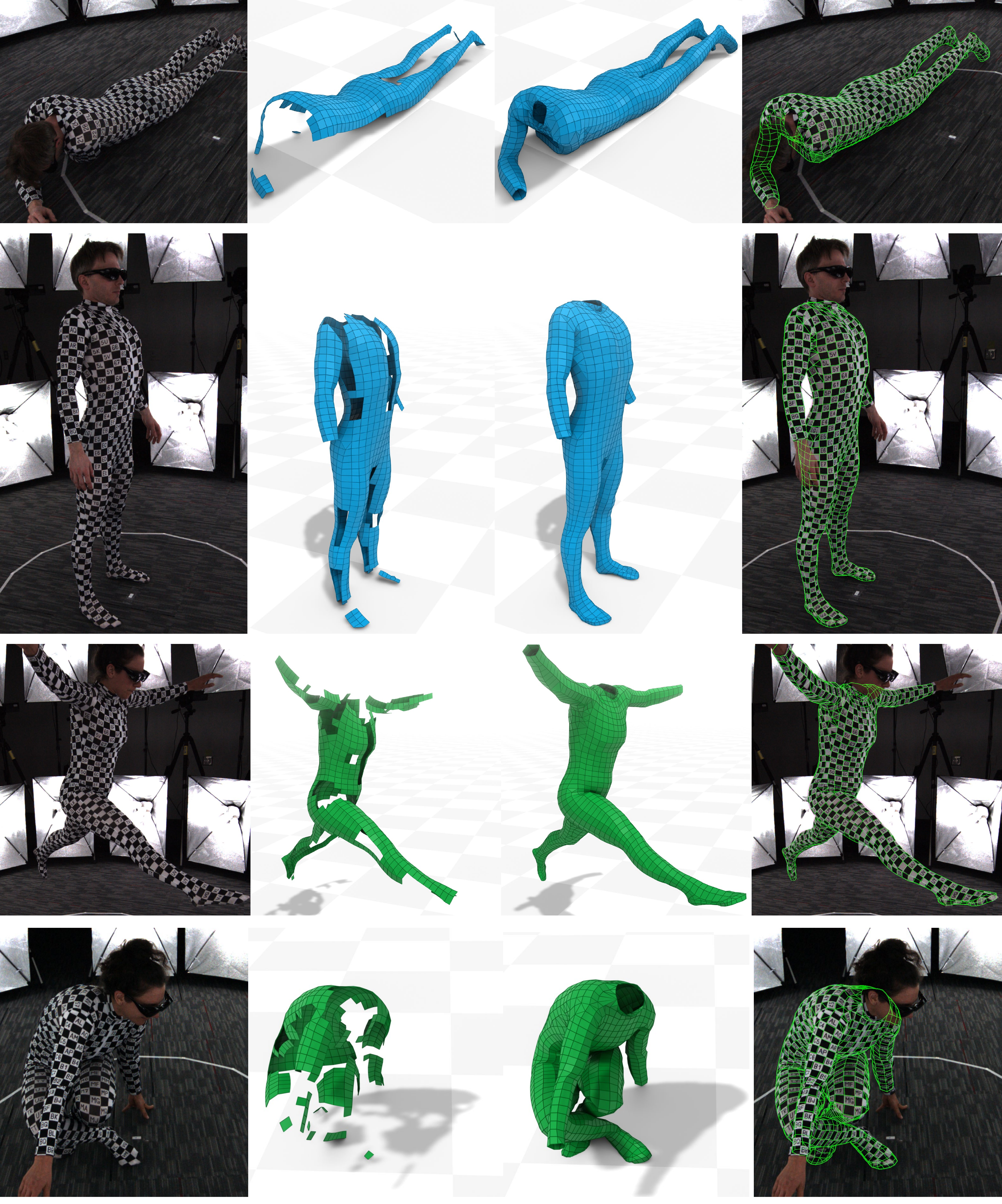}
    \caption{
       Results in challenging poses. From the left to right: 1) input images, 2) raw 3D reconstructions with holes, 3) our final meshes after interpolating non-articulated displacements, and 4) wireframe rendering of the final meshes overlaid with the original images.
    }
    \label{fig:Qualitative1}
\end{figure*}

To capture moving bodies with high detail, the DFAUST approach \cite{bogo2017dynamic} starts by geometrically registering a template body model to 3D scans \cite{hirshberg2012coregistration,pons2015dyna} and then uses colored patterns on the skin to obtain high-accuracy temporal correspondences via optical flow. These colored patterns serve a similar purpose as the checkerboard-like corners on our suit, i.e., they enable precise localization of points on the surface of the body. The key difference of our approach is that our suit contains also unique two-letter codes adjacent to each corner, allowing us to label the corners directly by recognizing the codes. This is not possible with the DFAUST's patterns, because they are self-similar, created by applying colored stamps to the skin. Instead, the DFAUST approach relies on the initial geometric registration and temporal tracking, which can suffer from error accumulation and may lead to incorrect local minima in more challenging poses or fast motions. The DFAUST dataset contains a variety of high-detailed human body animations, but is restricted to upright standing-type motions. In contrast, we demonstrate \ACChange{captures} of a wider variety of motions, including gymnastics exercises, yoga poses or rolling on the ground, see Fig. \ref{fig:Qualitative1} \ACChange{and our accompanying video and data}.


Our new motion capture method was enabled by recent advances in deep learning and high-resolution camera sensors. The key idea is to use a new type of motion capture suit with special fiducial markers, consisting of checkerboard-like corners for precise localization and two-letter codes for unique labeling. Our localization and labeling process is very robust, because it does not rely on temporal tracking or any type of body model; in fact, our approach succeeds even if only a small part of the body is visible in the image, e.g., the zoom-ins in the right part of Fig. \ref{fig:QualitativeResultsRecon}. A similar advantage exists also in the temporal domain. Because our localization and labeling approach can process each image independently, there are no issues due to occlusions and dis-occlusions which complicate traditional temporal tracking in both marker-based as well as marker-less methods.


Even though the automatic localization and labeling  \ACChange{are} very robust, achieving this functionality is non-trivial, because our methods need to be robust against significant stretching of the tight-fitting suit as well as projective distortion. Fortunately, checkerboard-like corners remain to be checkerboard-like even despite significant stretching of the suit. 
\ACChange{We apply three convolutional neural networks combined with geometric algorithms.}
The first of our convolutional neural networks (CNNs) is a corner detector which localizes all of the corners in an input image (4000 $\times$ 2160). To rectify the distortion of the two-letter codes inside the white squares, we connect candidate four-tuples of corners into quadrilaterals (quads) and apply homography transformation which rectifies both suit stretching as well as projective distortion. Another CNN called \Rejector{} then performs quality control and legible and upright-oriented codes. The remaining codes are passed to \RecogNet{} which reads the characters in the two-letter code. Because the orientation of our codes is unique (we avoided symmetric symbols such as ``O'' or ``I'' as well as ambiguous pairs like ``6'' and ``9''), recognition of the code allows us to uniquely label each of the adjacent corners, see Fig. \ref{fig:SuitDesign}c.

The labels of our corners establish correspondences both in time and in space, i.e., between individual cameras in our multi-view system, which means that we can easily triangulate the 2D corner locations into 3D points. However, the 3D reconstructed (triangulated) points will inevitably miss observations
due to self-occlusions and the limited number of our cameras, see Fig.~\ref{fig:teaster}c.
To fill (interpolate) these missing observations, we start by fitting the STAR model \cite{STAR:2020} and then refine it for each of our actors using a point trajectory from a calisthenics-type motion sequence captured using our method. This refinement ensures that we obtain the best possible low-dimensional model for each of our actors, since high quality is our main objective. We use this refined body model to interpolate the missing corners in the rest pose, resulting in the final mesh without any holes, see Fig. \ref{fig:teaster}d.

Our goal was to make each two-letter code as small as possible, so we can recover the highest possible number of points on the body. 
We created our special capture suits in two sizes, one ``medium'' (with 1487 corners) and one ``small'' (with 1119 corners) and we captured three actors: one male and two females (two of the actors used the ``medium'' suit). 

We evaluated both the geometric accuracy through reprojection error of 3D reconstruction, 
and the quality of the temporal correspondences by computing the optical flow between the synthetic image and the real image.
Results show 99\% of our reconstructed points has a reprojection error of less than 1.01 pixels, and 95\% of the pixels on the optical flow have a optical flow norm of less than 1.2 pixels.
In our camera setup, 1 pixel approximately converts to 1mm on a person 2 meters away from the camera.
\vspace{-5mm}
\paragraph{Contributions:}
\begin {enumerate*} 
\item We propose a new method to measure 3D marker locations at each frame and automatically infer corresponding \textbf{marker labels}. This is achieved without any priors on human body shapes or kinematics, which means that our data are ``raw measurements'', immune to any type of modeling or inductive bias. \\
\item We introduce a novel type of fiducial markers and capture suit, which enables marker localization and unique labeling using only local image patches. Our approach does not utilize temporal tracking, which makes it robust to marker dis-occlusions and also invites parallel processing, because each frame can \ACChange{be} processed independently. \\
\item All results in this paper were obtained using an experimental multi-camera system utilizing 16 commodity RGB cameras and passive lighting. 
High-end multi-camera systems based on machine vision cameras such as those built by Facebook \cite{lombardi2018deep} or Google \cite{guo2019relightables} require significant hardware investments and engineering expertise. In contrast, our system is easy to build from inexpensive \ACChange{off-the-shelf} parts. We provide our data as supplemental material and we invite individual researchers, independent studios or makers to replicate our setup and capture new actors and motions.
\end {enumerate*} 

\vspace{-3mm}
\section{Related Work}
Optical systems based on reflective markers \cite{menache2000understanding} are the most widely used approaches to capture \ACChange{the} human body. While typically only sparse marker sets are used, \cite{park2006capturing, park2008data} pushed the resolution of reflective markers based system up to 350 markers to capture the detailed skin deformation. However, difficulties in marker labeling \cite{song2016fast} complicate further increases of resolution by adding even more markers. Recent work \ACChange{utilizes} self-similarity analysis \cite{aristidou2018self} and deep learning \cite{holden2018robust, han2018online} to reduce the expensive manual clean-up in marker labeling procedure. An alternative to the classical reflective markers is the use of colored cloth, enabling \ACChange{the} capture of certain types of garments \cite{white2007capturing, scholz2005garment} or hand tracking using colored gloves \cite{wang2009real}. \looseness=-1

Early work in markerless motion capture \cite{gavrila1996tracking} and \cite{bregler2004twist} inferred human poses directly from 2D images or videos.
\cite{kehl2006markerless} integrates multiple image cues such as edges, color information and volumetric reconstruction to achieve higher accuracy.
\cite{brox2009combined} tracks a 3D human body on 2D image by combining image segments, optical flows and SIFT features \cite{lowe1999object}.
\cite{de2008performance} deforms laser-scan of the tracked subject under the constraints of multi-view videos to capture spatio-temporally coherent body deformation and textural surface appearance of the actors.
Silhouettes \cite{vlasic2008articulated, liu2013markerless, sand2003continuous, starck2007surface} or visual hulls \cite{corazza2010markerless} can be used to obtain more detailed human body deformations. 
\cite{gall2010optimization} introduce a multi-layer framework that combines stochastic optimization, filtering, and local optimization to tack 3D human poses. 
\cite{stoll2011fast} model the human body via a sums of Gaussians, representing both shape and appearance of the captured actors.

Deep learning enabled estimation of 2D or 3D human poses from multi-view \cite{pavlakos2017harvesting} or monocular multi-person images \cite{newell2016stacked, pishchulin2016deepcut, raaj2019efficient, wei2016convolutional},
more recently also with hands and faces \cite{cao2017realtime, cao2018openpose,hidalgo2019single}. 
3D pose or even dense 3D surface of the human body can also be predicted from a single image \cite{mehta2017vnect, choutas2020monocular, xiang2019monocular, guler2018densepose, xu2019denserac}. 
Morphable human models can be learned from multi-person datasets \cite{pons2015dyna, robinette2002civilian}. Models such as SCAPE \cite{anguelov2005scape}, SMPL \cite{loper2015smpl} and STAR \cite{STAR:2020} focus on the body, while models such as Adam \cite{joo2018total} and SMPL-X \cite{pavlakos2019expressive} also include the face and the hands. Focusing on high-quality rendering rather than geometry, \cite{guo2019relightables, Meka:2020} proposed methods for photo-realistic relighting of moving humans, including clothing and accessories such as backpacks.

\begin{figure}
    \setlength\belowcaptionskip{-3ex}
    \centering
    \includegraphics[width=0.9\columnwidth]{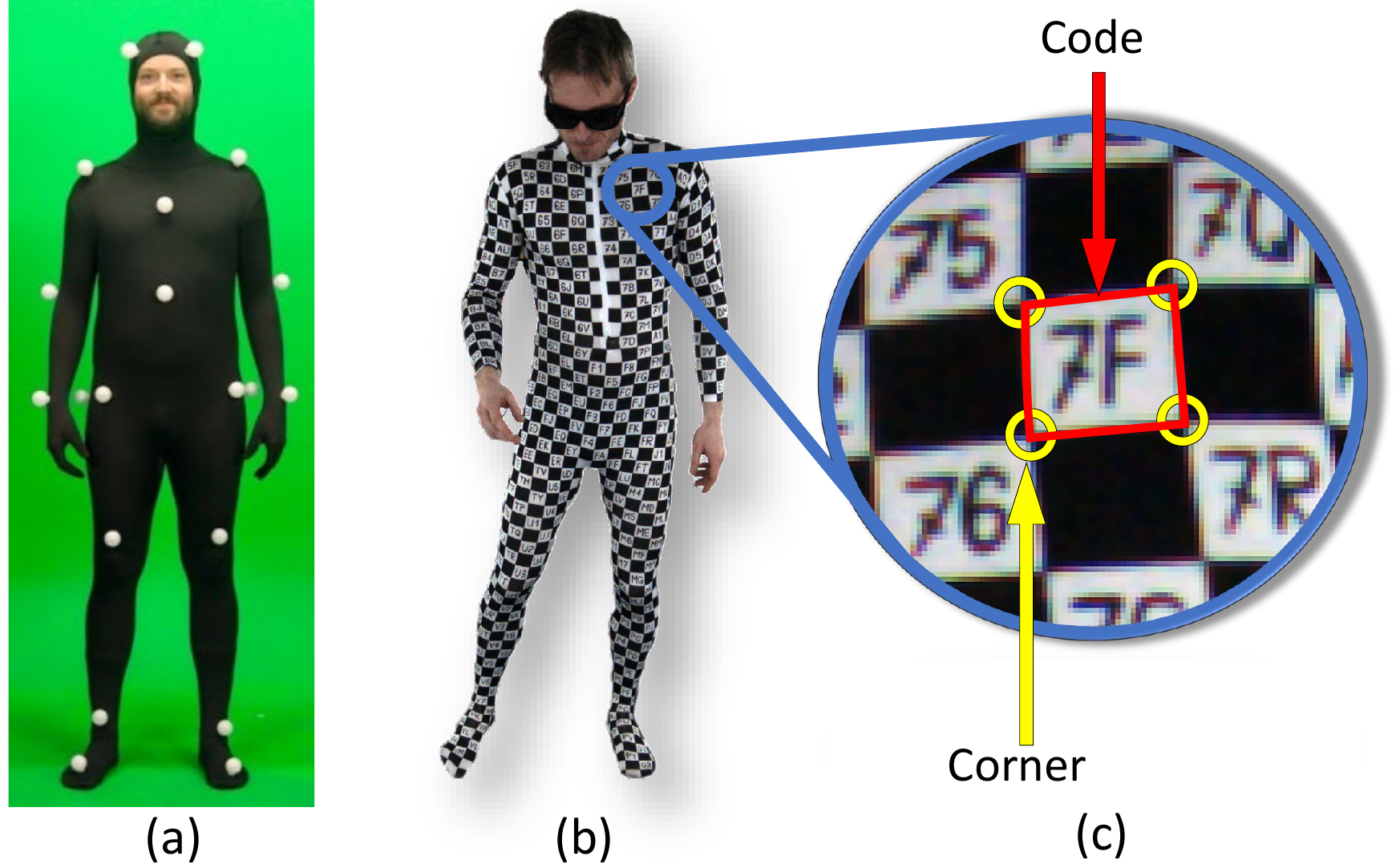}
    \caption{(a) A classical motion capture suit with reflective markers. (b) The design of our motion capture suit with fiducial markers. (c) Each or our markers consists of checkerboard-like corners and codes.}
    \label{fig:SuitDesign}
\end{figure}

The idea of a motion capture suit with special texture is related to fiducial markers used e.g., in robotics or augmented reality, such as ARTag \cite{fiala2005artag}, AprilTag \cite{olson2011apriltag, wang2016apriltag}, ArUco \cite{garrido2014automatic} and many others, but these fiducial markers are typically assumed to be non-deforming. \ACChange{They are also not easy to read for humans, which would complicate their annotations}. 
The localization of our fiducial marker is related to corner detection. Many corner detection methods have been developed to meet different use-case scenarios. For localizing the corners, there are methods that are designed to detect general  corners features that naturally exists in nature, like \cite{detone2018superpoint, rosten2006machine}, Another class of corner detectors focuses on rigid calibration checkerboards \cite{bennett2014chess, hu2019deep, chen2018ccdn, donne2016mate}, particularly useful in camera calibration. Because these methods assume the checkerboard pattern to be rigid, they will not work on our checkerboard-like suit which can have significant deformations (Fig. \ref{fig:SuitDesign}b). The code recognition component of our method is related to text recognition. As discussed in \cite{long2020scene}, the text recognizer generally performs poorly on text with large spatial transformations. One possible solution is based on generating region proposals \cite{ma2018arbitrary,jaderberg2014deep} to rectify the spatial transformation.

High-resolution temporal correspondences can be obtained by registering a template mesh to  RGB-D images or 3D scans. The registration can be based on solely geometric information \cite{allain2015efficient, li2009robust}, or combined with RGB images to align \cite{bogo2015detailed} to reduce the tangential sliding. Model-less approaches are also possible \cite{dou20153d, newcombe2015dynamicfusion, collet2015high}.
Those methods focus on registering sequential motions frame to frame, with the assumption of small displacements between subsequent frames. Therefore, they can suffer from error accumulation, resulting in drift over time \cite{casas20124d}. 
Aligning non-sequential motions is also possible \cite{huang2011global,tung2010dynamic}, but it is challenging to establish correspondence between very different poses \cite{boukhayma2016eigen, prada2016motion}.
Deformation models can be trained from 3D scans \cite{allen2003space,anguelov2005scape}, with non-rigid scan registration being the technical challenge \cite{hirshberg2012coregistration}. 

Similarly to our new motion capture suit, the FAUST \cite{bogo2014faust} and DFAUST \cite{bogo2017dynamic} methods paint high-frequency colored patterns directly to the skin. We chose to work with a suit because putting it on and off is easy and fast compared to applying colored stamps and washing them off after the capture session. Even though we only experimented with basic tight-fitting suits in this paper, future improvements such as adhesive suits or non-permanent tattoos are possible, see Section \ref{sec:Limitation}. Our capture system is significantly simpler and less expensive: we use only on 16 standard (RGB) cameras with passive uniform lights, while \cite{bogo2014faust,bogo2017dynamic} used 22 pairs of stereo cameras, 22 RGB cameras and 34 speckle projectors (active light). Perhaps more important are the technical differences between our approach and DFAUST, in particular the fact that our codes are unique as opposed to the self-repetitive patterns used in FAUST and DFAUST. Rather than creating a dataset, our goal was to create a universal and practical method to enable future research on advanced human body modeling and its applications in areas ranging from graphics to sports medicine.

\vspace{-3mm}
\section{3D Reconstruction of Labeled Points}
\label{sec:CaptureSystem}
\begin{figure}
    \centering
    \setlength\belowcaptionskip{-3ex}
    \includegraphics[width=0.9\columnwidth]{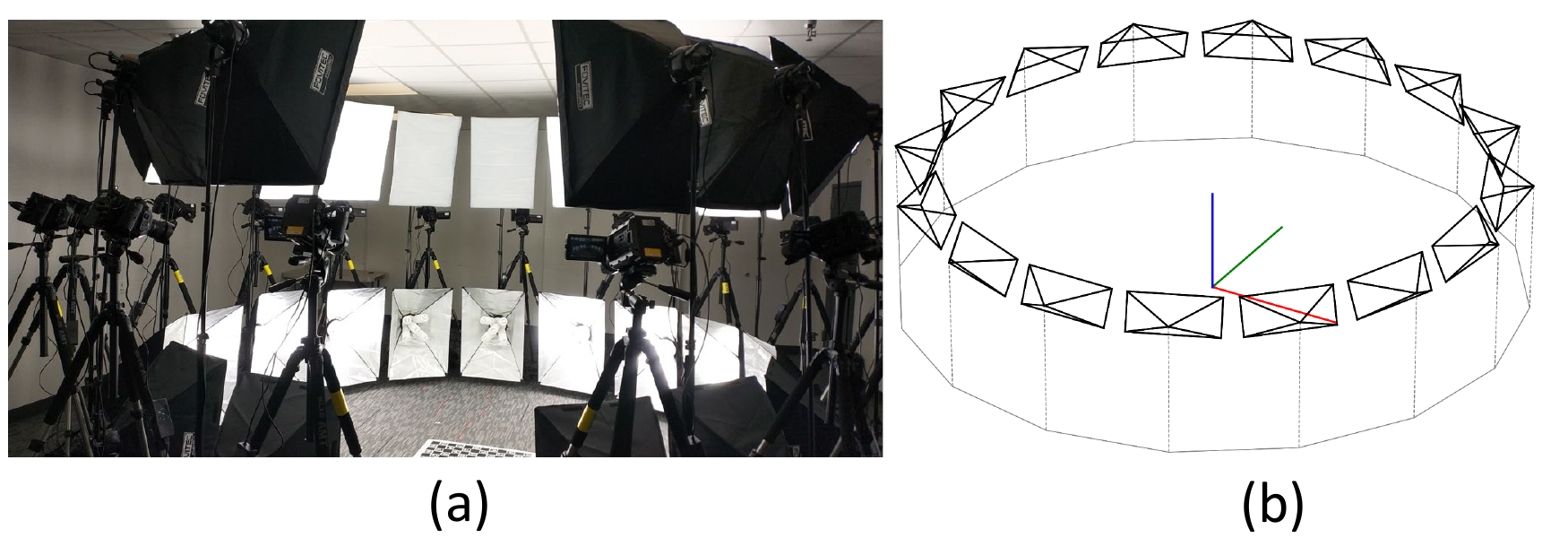}
    \caption{(a) A photo of our 16 camera setup. (b) The cameras form a circle surrounding the capture volume.}
    \label{fig:System}
\end{figure}
\begin{figure*}
    \centering
    \setlength\belowcaptionskip{-3ex}
    \includegraphics[width=0.95\textwidth]{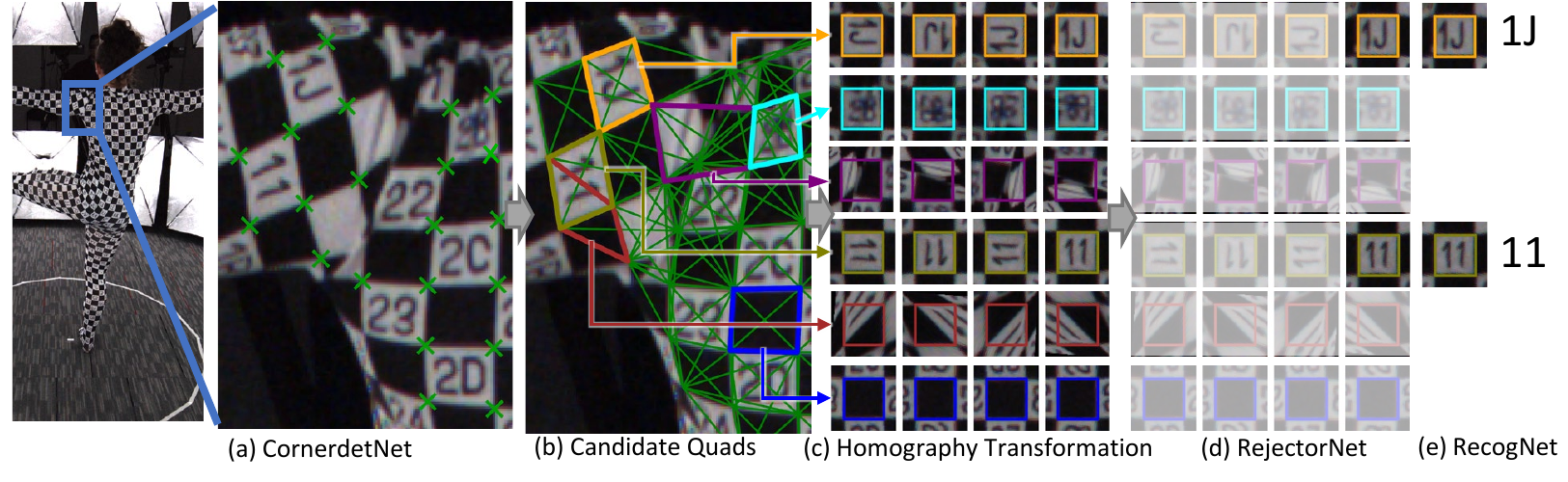}
    \caption{Corner localization and labeling pipeline: (a) \Cornerdet{} CNN detects and localizes our suit corners; (b) Four-tuples of corners are connected into candidate quads; (c) homography transformations using all four possible orientations (we selected a few quads and outlined them with different colors for illustration purposes); (d) \Rejector{} CNN: a binary classifier accepting only valid white squares with upright oriented two-letter codes; (e) \RecogNet{} CNN recognizes the two characters in the valid codes.}
    \label{fig:CNNPipeline}
\end{figure*}
\paragraph{Suit.} To create our special motion capture suit, we started by purchasing a tight-fitting unitard, originally intended for dance or performing arts. Fortuitously, one of the manufacturer-provided patterns was precisely the black-and-white checkerboard texture reminiscent of computer vision calibration boards (in fact this provided some of the original inspiration for this project). We purchased two suits, one ``medium'' and one ``small'' and augmented them by writing codes into the white squares using a marker pen. The medium suit contains 1487 corners and 625 two-letter codes; the small suit has 1119 corners and 456 codes. 
For our two-letter codes, we only used symbols whose upright orientation is unique and non-ambiguous, specifically: ``1234567ABCDEFGJKLMPQRTUVY''.
\begin{figure}
    \centering
    \setlength\belowcaptionskip{-3ex}
    \includegraphics[width=\columnwidth]{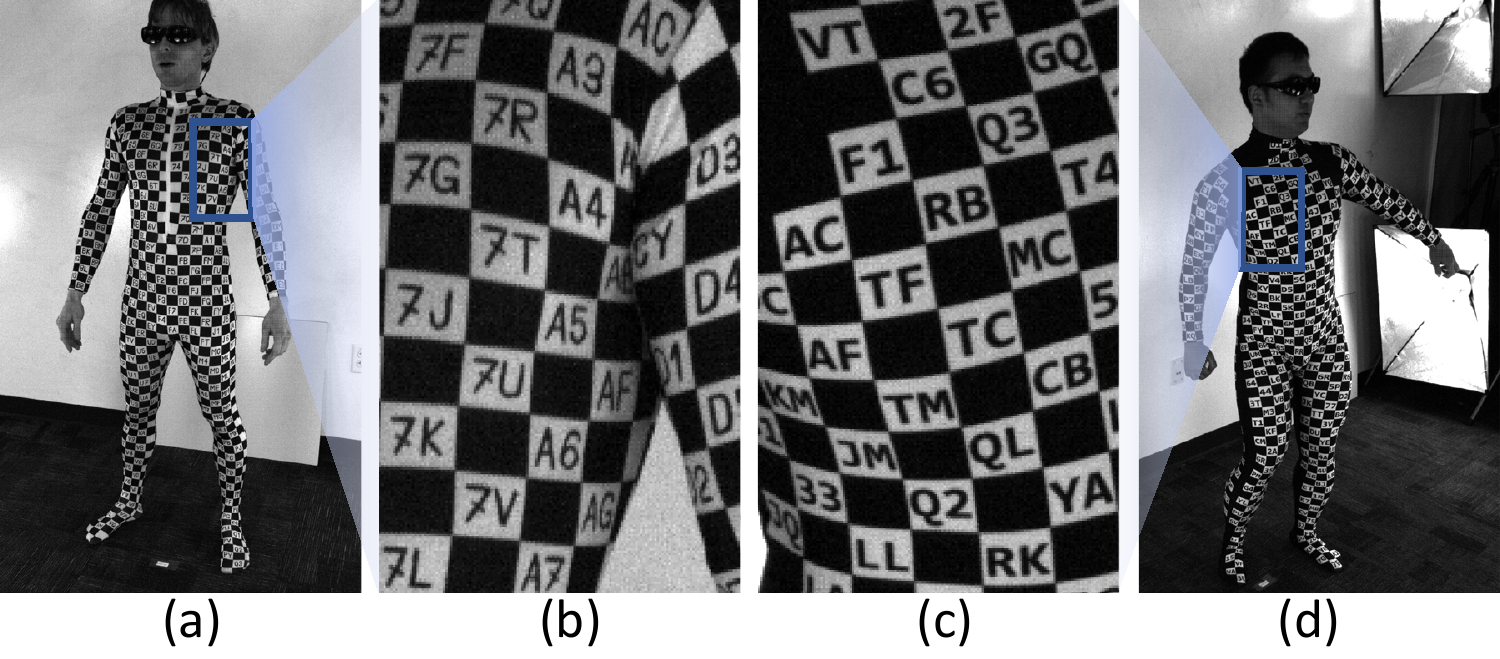}
    \caption{Comparison of (a, b) hand-drawn and (c, d) printed suits.}
    \label{fig:SuitComrHanddrawnPrinted}
\end{figure}

An alternative to hand-drawn suits is to use a cloth printing service, capable of printing an arbitrary high-resolution image on textile. However, this approach is more complex because it requires us to design sewing patterns, \Rv{ship} them to the printing service, then \Rv{cut} and \Rv{sew} the printed textile into a suit. We have explored this approach, using a computer font that is quite different from our handwriting, see Fig. \ref{fig:SuitComrHanddrawnPrinted}c,d. This experimental suit contains 1473 corners and 618 two-letter codes, but it contains large black areas due to suboptimal sewing patterns.

\paragraph{Camera system.} Our multicamera setup contains 16 standard (RGB) cameras arranged into a circle surrounding the capture volume (Fig. \ref{fig:System}b). Each camera captures 4000$\times$2160 images at 30 FPS in RAW format. The camera shutters are synchronized via genlock with negligible synchronization error, which means that human motion is captured as if ``frozen'' in time. Surrounding the capture volume, we positioned 32 softboxes that generate uniform diffuse light. 
The bright light allows us to use a small aperture and very fast $0.5ms$ shutter speeds, guaranteeing sharp images even with the fastest human motions. The cameras are calibrated by waving a traditional calibration checkerboard in from of them. 
The intrinsic and extrinsic camera parameters are calibrated using the well-established method \cite{zhang2000flexible} for which we use OpenCV's checkerboard corner detector for rigid calibration boards \cite{opencv_library}. Next, the camera parameters and the 3D checkerboard corner positions in the world coordinates are further refined using bundle adjustment \cite{triggs1999bundle}. We use the Levenberg-Marquardt algorithm and the Ceres library \ACChange{\cite{agarwal2012ceres}}. 

\vspace{-3mm}
\paragraph{Image processing pipeline.} The calibrated cameras \ACChange{generate} sequences of images, which are processed by our pipeline outlined in Fig. \ref{fig:CNNPipeline}. We start by detecting checkerboard-like corners in the input image  with sub-pixel accuracy (Fig. \ref{fig:CNNPipeline}a, Section \ref{subsec:CornerdetNet}). Next, we need to uniquely label the detected corners by recognizing the adjacent two-letter codes. Because the codes are written in the white squares surrounded by four corners, we generate candidate quadrilaterals (quads) by connecting four-tuples of corners. Only a few four-tuples of corners correspond to the white squares, but it is okay to generate a quad that does not correspond to a white square, because it will be discarded later; hence we use the term ``candidate quads'', see Fig. \ref{fig:CNNPipeline}b. Since the quads are generated by connecting four corners, we naturally have correspondences between the corners and the quads.
The candidate quads are rectified by mapping them into a regular square using homography (Fig. \ref{fig:CNNPipeline}c, Section \ref{subsec:QuadProposal}) to remove suit stretching and perspective distortion, and then passed as input to \Rejector{} which performs quality control and checks whether the quad actually corresponds to a white square with a code (Fig. \ref{fig:CNNPipeline}d). The \Rejector{} also ensures \ACChange{the} correct upright orientation of the code. The images accepted by \Rejector{} are then passed to \RecogNet{}, which reads the two-letter code that finally enables us to uniquely label each corner (Fig. \ref{fig:CNNPipeline}e).

We would like to point out that our method is local by design, i.e., each stage of the pipeline works with small patches of the input image. This gives us a several advantages:
\begin {enumerate*} [label=\itshape\alph*\upshape)]
\item Our method is capable of extracting reliable geometric information of the human body and - crucially - correspondences even from a small patch of the suit. This makes our method very robust to occlusions or partial views of the human body e.g. due to zoomed-in cameras.
\item By decomposing the suit into small quads and  undistorting them using homography, we can counteract much of the projective distortions and \ACChange{suit stretching} (see Fig.\ref{fig:CNNPipeline}c), simplifying the learning task. 
\item The CNN quad classifier includes a quality control mechanism, rejecting white squares with dubious quality and further improving the robustness of our method. 
\end {enumerate*} 

\vspace{-5mm}
\subsection{Corner Detector}
\label{subsec:CornerdetNet}
The corner detector's task is to detect and localize all checkerboard-like corners in the input image. This task is non-trivial because there are corner-like features in the background, the suit stretches along with the skin and there are significant lighting variations.

Our corners have two key properties:  
\begin {enumerate*} [label=\itshape\alph*\upshape)]
\item The corners are sparsely and approximately uniformly distributed on the suit; 
\item The corners are defined locally, i.e., a small image patch is enough to identify and localize a corner.
\end {enumerate*} 
We divide our input image into a grid of $8\times8$ cells  with the assumption that there could be at most one checkerboard-like corner in each $8\times8$ cell, and apply \Cornerdet{} CNN to detect and localize a checkerboard-like corner from each cell separately. The design of \Cornerdet{} is inspired by single shot detectors  \cite{redmon2017yolo9000, liu2016ssd}, which perform prediction and localization simultaneously. 
Please see Supplementary Material 
A.1 for more details about \Cornerdet{}.

\subsection{Corner Labeling and 3D Reconstruction}
 \label{subsec:QuadProposal} 

\begin{figure}
    \centering
    \setlength\belowcaptionskip{-3ex}
    \includegraphics[width=0.9\columnwidth]{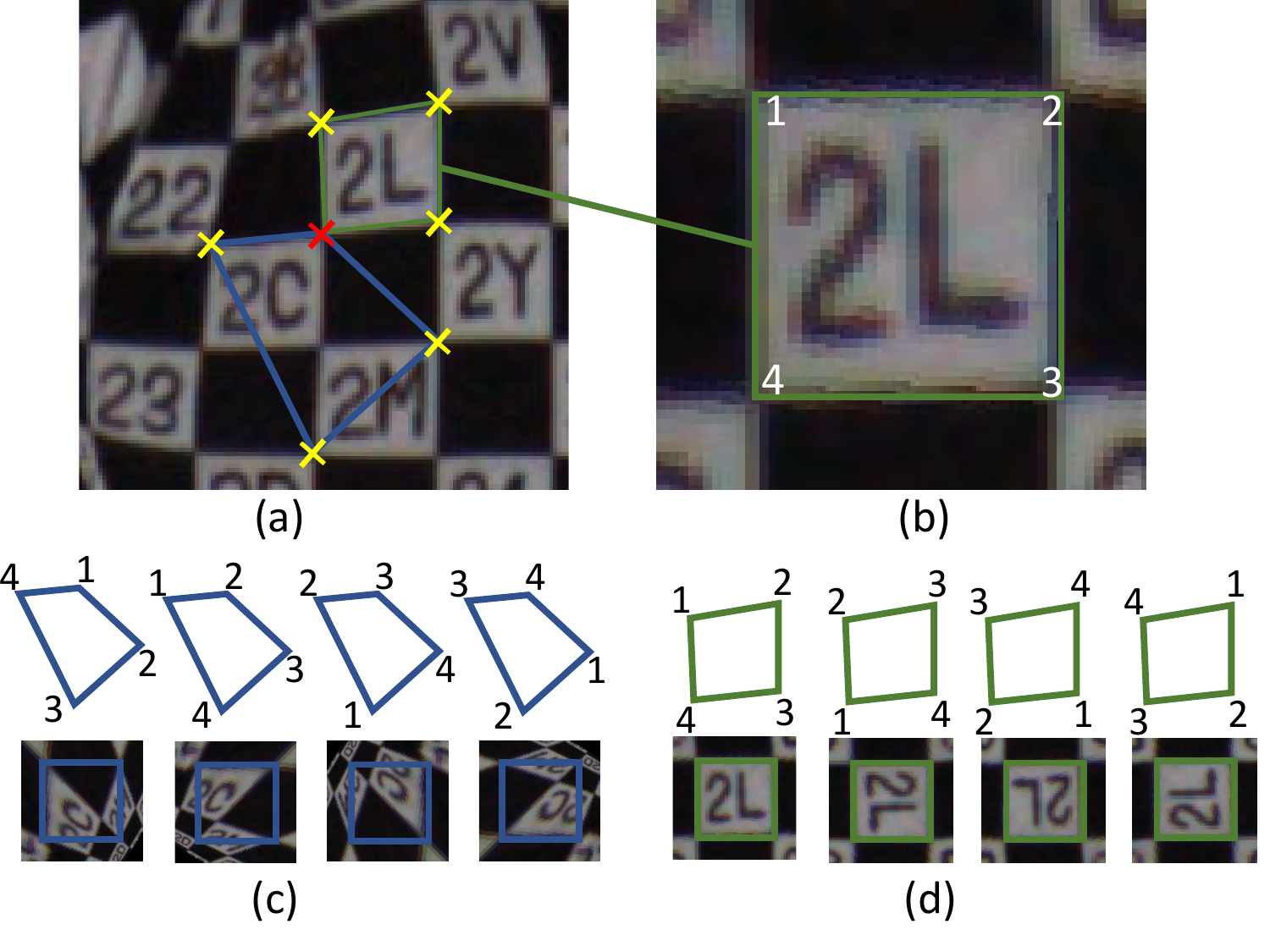}
    \caption{Quad generation: (a) For a given corner (red), we find three other corners (yellow) within a bounding box and compute their convex hull to generate a candidate quad. Here we visualize two example candidate quads (green and blue); (b) Our corner numbering convention with respect to upright code orientation; (c) and (d): The four possible orientations of the candidate quads, and their corresponding homography transformations. (c) is an invalid candidate, but the first orientation in (d) is valid.
    }
    \label{fig:CandidateQuadsGeneration}
\end{figure}

At this point we have detected typically several hundreds of corners in each input image. The next step is to read the codes and link them to the corners, which will give us \ACChange{a unique label} for each corner. As shown in Fig. \ref{fig:teaster}, the deformations of the suit and its codes can be significant, including not only projective transformations, but also stretching and shearing because the tight-fitting suit is highly elastic. 
First we have to detect the white squares with two letter codes. We know that each such white square is surrounded by four corners. Therefore, we generate quadrilaterals (quads) by connecting four-tuples of corners. In theory we could connect any four-tuples of corners into a quad, but in practice we can immediately discard concave quads (which do not correspond to correct sequences of corners) or quads that would cover too few or too many pixels (which would make it impossible to contain a legible code).
We call the resulting quads ``candidate quads'', because they may - but are not guaranteed \ACChange{to} - contain a correct two-letter code. We transform the four corners of each candidate quad to a standardized square using a homography transformation to simplify subsequent processing. The standardized square is a $64\times 64$ pixel image with $20$ pixel margin on each side, i.e., $104\times 104$ total, see Fig. \ref{fig:CandidateQuadsGeneration}. The $20$ pixel margin allows the \Rejector{} to detect errors stemming from incorrect corner detections. Since we do not know the correct upright orientation of our two-letter code yet, we generate all four possible orientations, see Fig. \ref{fig:CandidateQuadsGeneration}c,d.
Fig.~\ref{fig:CandidateQuadsGeneration}c shows an example of an invalid quad and Fig.~\ref{fig:CandidateQuadsGeneration}d demonstrates a valid one.
\paragraph{Quad classifiers.} We trained two quad classifiers, \Rejector{} and \RecogNet{}. 
\Rejector{} \ACChange{is a binary classifier predicting} whether a candidate quad is valid, i.e., whether the four corners are at the correct locations and their order is correct relative to the upright code orientation, see Fig. \ref{fig:CandidateQuadsGeneration}b. Also, the white square \ACChange{surrounded by a valid quad needs to contain a clearly legible code}.  Invalid quads are discarded, and the valid ones are passed to \RecogNet{} which reads the codes, such as ``2L'' in Fig.~\ref{fig:CandidateQuadsGeneration}b.
\RecogNet{} is a multi-class classifier with two heads, one for each character \ACChange{ of the two letter code}. 
The details of candidate quad generation and quad classifiers can be found in Supplementary Material A.2.
We use standard cross entropy losses to train those classifiers.
\ACChange{The training of our CNNs is discussed in Section \ref{sec:Results}.}

\paragraph{Corner labeling and 3D Reconstruction.} At this point, the two-letter codes of the valid quads have been recognized, including their upright orientation. The next step is to uniquely label each corner. 
We define a labeling function $l(code, i_q)$  which maps a two-letter $code$ and corner index $i_q\in\{1,2,3,4\}$ \ACChange{(see Fig. \ref{fig:CandidateQuadsGeneration}b)} to \ACChange{an integer which represents a} unique corner ID. The unique corner \ACChange{IDs} are defined for each suit. Many corners have \textit{two} two-letter codes adjacent to them. If both of the two-letter codes are visible, we can leverage this fact as a redundancy check, detecting potential errors of \RecogNet{}.
Given unique corner IDs, we can convert corresponding 2D corners in more than two views into labeled 3D points. 
Let $\mathcal{C}_i$ be the set of cameras that see corner $i$, $k \in \mathcal{C}_i$ is a camera that sees corner $i$, $\mathbf{c}_i^k \in \mathbb{R}^2$ be the corner $i$'s location in image coordinate system of camera $k$, and $f^k:\mathbb{R}^3 \to \mathbb{R}^2 $ is the projection function of camera $k$. We computed 3D reconstructed corner $\mathbf{p}_i$ by minimizing the reprojection error:
\begin{equation}
\label{eq:reconstruction}
\mathbf{p}_i = \underset{\mathbf{p}_i\in \mathbb{R}^3}{\arg\min} \sum_{k\in \mathcal{C}_i } ||  f^k(\mathbf{p}_i)-\mathbf{c}_i^k||^2_2 
\end{equation}
This is a non-linear least square optimization problem; we compute an initial guess of $\mathbf{p}_i$ using the Linear-LS method \cite{hartley1997triangulation} and optimize it using non-linear least squares \ACChange{solver \cite{agarwal2012ceres}.}
\begin{figure}
    \centering
    \includegraphics[width=\columnwidth]{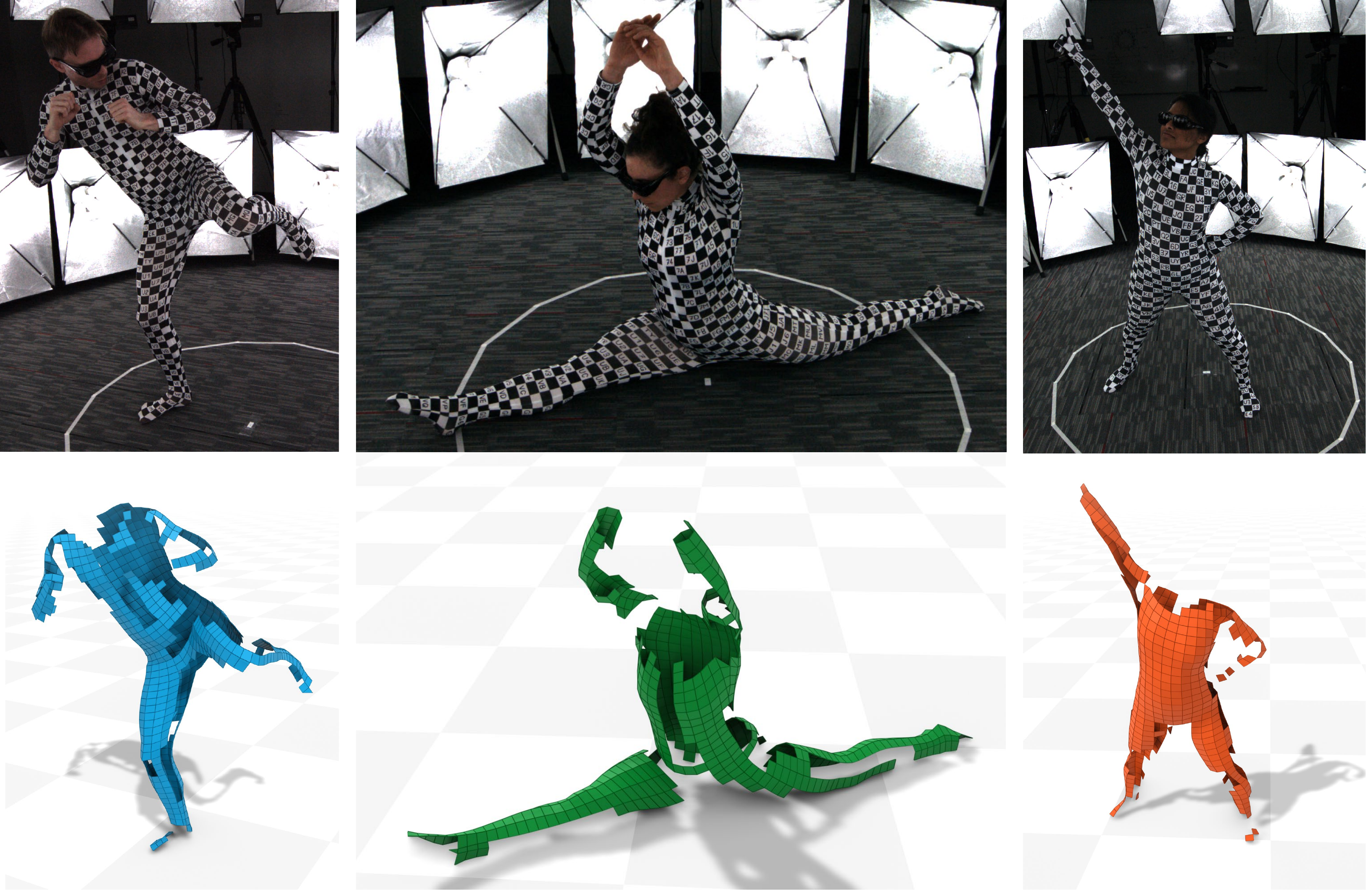}
    \caption{
         Example input images (top) from our multi-camera system and the corresponding raw 3D reconstructions (bottom). The reconstructed 3D points are meshed according to the patterns in our suits.
    }
    \label{fig:RawReconstruction}
\end{figure}

\paragraph{Error Filtering}
The label consistency check discussed above works only if two adjacent two-letter codes are present in the suit and visible in the images. If this is not the case, a corner can be assigned  the wrong label if \Rejector{} or \RecogNet{} make a mistake. This kind of labeling errors will typically result in non-sensical correspondences with large reprojection errors, which we detect and correct by a RANSAC-type method discussed below. Specifically, for a corner with label $i$, let $\mathcal{C}_i$ be the set of the cameras that claim to see this corner. We assume that outliers in $\mathcal{C}_i$, i.e., the cameras that mislabeled corner $i$, should only be a minority.
We iterate over all pairs of cameras $(j,k)$ in  $\mathcal{C}_i$, and 3D reconstruct the corner $i$ from each pair. Among all of the pairs, we pick the 3D reconstruction that has the lowest reprojection error averaged over all cameras \ACChange{in} $\mathcal{C}_i$ and assume this is the correct 3D location $\mathbf{p}_i$.
Next, we analyze the reprojection errors of $\mathbf{p}_i$ into all of the cameras $\mathcal{C}_i$. 
The reprojection error should be low in cameras with correct labeling, but high if there was a labeling error. We use the $1.5\times IQR$ (interquartile range) rule \cite{upton1996understanding} to detect the outliers in terms of reprojections errors. We re-compute the triangulation of $\mathbf{p}_i$ after removing the outliers from the cameras $\mathcal{C}_i$. This RANSAC-type outlier filter does not work when there are only two cameras that see one corner. Therefore, we additionally discard reconstructed corners with \ACChange{an} average reprojection error larger than $1.5$ pixels. These tests are designed to be conservative, because mistakenly discarded points are not a major problem, just missing observations which can be inpainted as discussed in Section~\ref{sec:HoleFilling}.


\vspace{-3mm}
\section{Data Acquisition and Neural Network Training}
\label{sec:DataGen}
A key feature of our approach is that all of our networks are trained only on small image patches, e.g., see Fig. \ref{fig:CandidateQuadsGeneration}c,d. This allows our trained model to generalize to different suits,
capture environments, camera configurations and body poses that are not in the training set, because our local fiducial markers exhibit significantly less variability than images of full human poses. This is quite different from deep-learning based methods that perform global pose-prediction, looking for the body as a whole.
The training of our networks does not require large training sets. We have prepared our training data ourselves, without the use of any external annotation services or existing data sets.
\begin{figure}
    \centering
    \setlength\belowcaptionskip{-3ex}
    \includegraphics[width=0.9\columnwidth]{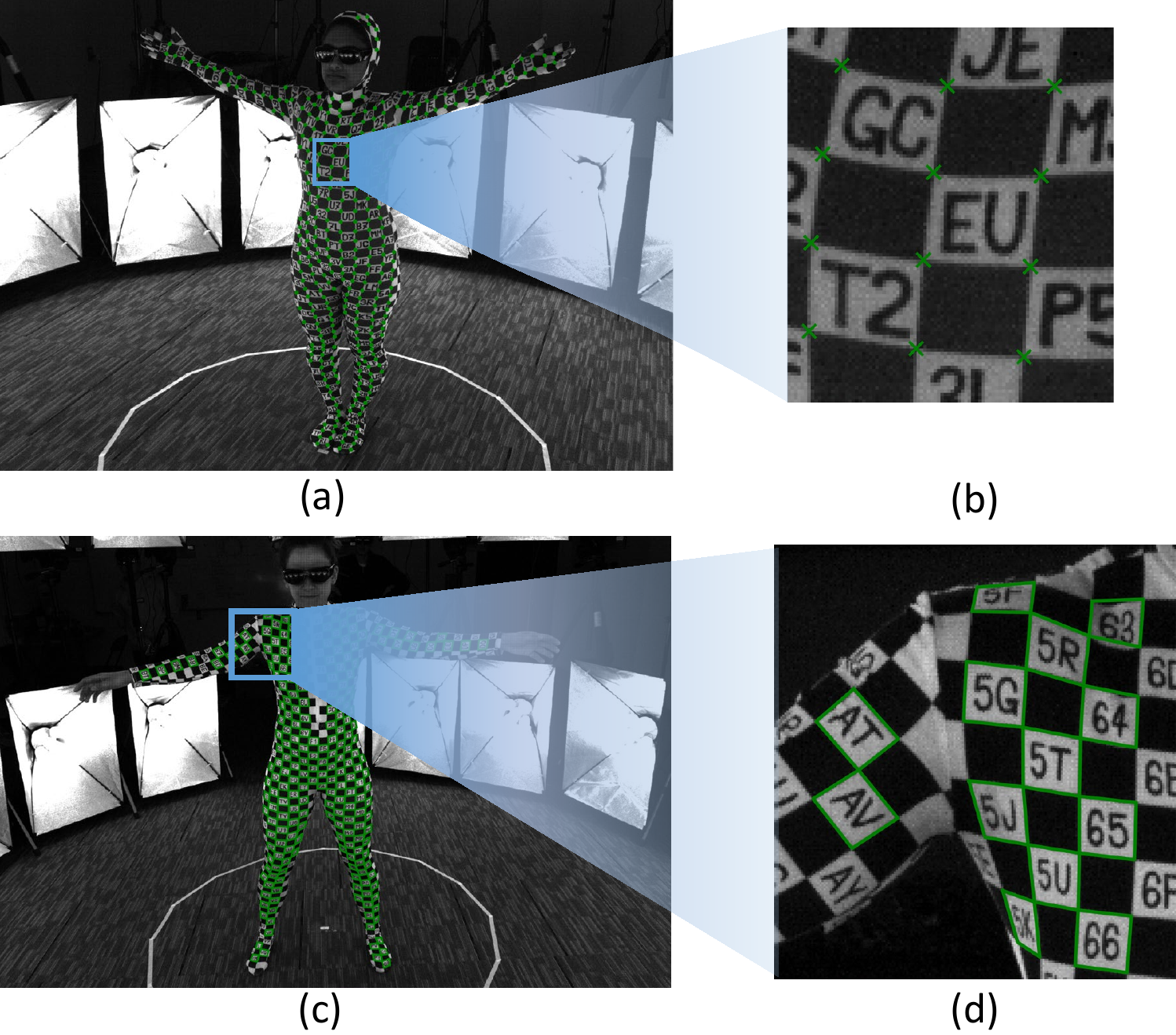}
    \caption{Two types of annotations we applied: (a, b): corner annoatation; (c, d): quad annotation.}
    \label{fig:DataAnnotationCornerAndQuad}
\end{figure}
Our dataset contains 28 manually annotated images, 24 of them are randomly selected from captures of our three actors wearing the  hand-drawn suits. We also captured a different (fourth) actor wearing the printed suit and annotated four images of it in order to evaluate our approach on printed characters as opposed to hand-drawn. \Rv{The data from the printed suit will only be used for testing the performance of CNNs.} For each image, we apply two types of annotations: corner annotation (\Fig{fig:DataAnnotationCornerAndQuad}a,b) and quad annotation (\Fig{fig:DataAnnotationCornerAndQuad}c,d). In the corner annotation, we manually annotate all of our checkerboard-like corners on the suit with sub-pixel accuracy. In the quad annotation, we manually connect the corners annotated in the previous step into quads. Specifically, we create quads that correspond to valid white squares with two-letter codes in the suit and the annotators also write down the code of each annotated quad. We ensure the quad vertices are in a clockwise order and start from the top-left corner, defined by the upright orientation of the code (see \Fig{fig:CandidateQuadsGeneration}b).
It takes about two person-hours to annotate one image.
These annotations are then automatically converted into training data for our networks using techniques described in \AppendixDataAug{}. 

We apply data augmentation to our training data through geometric and color space operations. We also generate synthetic training data by rendering textured human body models. The details can be found in \AppendixSynthData{}.
\section{Filling Missing Observations with Refined Body Model}
\label{sec:HoleFilling}

Our method for 3D reconstruction of labeled points will inevitably result in missing observations because the human body often self-occludes itself and is observed only by a limited number of cameras, see Fig. \ref{fig:RawReconstruction}. In this section, we propose a method to interpolate (inpaint) the missing corners. Even though we could use any existing multi-person human body model \cite{loper2015smpl, STAR:2020} for this purpose, we can achieve higher quality, because our pipeline gives us highly accurate measurements of the actor's body and its deformations. Therefore, instead of relying on previous statistical body shape models, we capture example motions of a given actor using our method and use this data to create a more precise \textit{refined body model}, i.e., a model with parameters refined for a specific person. 

Our body model has two types of parameters: shape parameters that are invariant in time, and pose parameters that change from frame to frame as the body moves. The shape parameters are only optimized during the model refinement process. After the body model refinement process is done, we fix the shape parameters and only allow the pose parameters to change. 
However, even after the refinement, the low-dimensional body model will not fit the 3D reconstructed corners exactly (Fig. \ref{fig:TrackingPointsCompletion}c).
We call the remaining residuals ``non-articulated displacements'', because they correspond to motion that is not well explained by the articulated body model. The non-articulated displacements arise due to breathing, muscle activations, flesh deformation, etc. Therefore, in addition to our refined body model we also interpolate the non-articulated displacements mapped to the rest pose via inverse skinning. The combination of the refined body model with the non-articualted displacement interpolation enables us to achieve high quality inpainting.

\subsection{Actor-Specific Model Optimization}

\newcommand{\restpose}[1]{\mathbf{\tilde{#1}}}

\label{subsec:bodyModel}
Our body model is based on linear blend skinning (LBS) \cite{magnenat1988joint}. 
Let $\mathbf{v}_i \in \mathbb{R}^4$ for $i = 1,2, \dots, N$ be the deformed vertices in homogeneous coordinates, where $N$ is the number of all the vertices on our body model.
We denote the skinning model as: $\mathbf{v}_i = D_i(\restpose{v}_i, \mathbf{J}, \mathbf{W}, \mathbf{\theta})$, where $\restpose{v}_i \in \restpose{V}=( \mathbf{\tilde{v}_1}, \dots, \mathbf{\tilde{v}}_N)$, 
 $\restpose{V}$ are rest pose vertex positions, $\mathbf{J}=(\mathbf{j_1}, \dots, \mathbf{j}_M)$ are joint positions and $M$ is the number of joints in our model; $\mathbf{W}\in \mathbb{R}^{N\times M}$ is the matrix of skinning weights and $\theta\in \mathbb{R}^{M\times 4}$ are joint rotations represented by quaternions. In summary, $\restpose{V}$, $\mathbf{J}$ and $\mathbf{W}$ are shape parameters (constant in time) and $\theta$ are pose parameters (varying in time).
The deformed vertices can be computed as:
\begin{equation}
\mathbf{v}_i = D_i(\restpose{v}_i, \mathbf{J}, \mathbf{W}, \theta)=  \sum_{j=1}^{M} w_{i,j} T_j(\theta, \mathbf{J}) \Rv{\restpose{v}_i}
\end{equation}
%
Note that here $\mathbf{v}_i, \restpose{v}_i \in \mathbb{R}^{4\times 1}$ are the deformed and rest pose vertex in homogeneous coordinates and $ \Rv{T_j}(\theta, \mathbf{J}) \in \mathbb{R}^{4\times 4}$ represents the transformation matrix of joint $j$.
In the following we will use homogeneous coordinates interchangably with their 3D Cartesian counterparts.
 
\paragraph{Initialization.} We initialize our body model by registering our corners to the STAR model \cite{STAR:2020}. We start by selecting a frame $f_{\mathrm{init}}$ in a rest-like pose where most corners are visible, and fit the STAR model to our labeled 3D points in $f_{\mathrm{init}}$ using a non-rigid ICP scheme which finds correspondences between our suit corners and the STAR model's mesh. The non-rigid ICP process is initialized by 10 to 20 hand picked correspondences between the STAR model and the 3D reconstructed corners.
During the ICP procedure, We optimize both pose and shape parameters of the STAR model and iteratively update correspondences by projecting each of our 3D reconstructed points to the closest triangle of the STAR model (the actual closest point is represented using barycentric coordinates).


In this stage, we have registered most of our corners to the STAR model, but we still need to add corners that were unobserved in frame $f_{\mathrm{init}}$. We can fit the STAR model to subsequent frames of our training motion using non-rigid ICP initialized by the registered corners instead of hand picked correspondences. 
These subsequent frames will reveal corners unobserved in the initial frame, which we register against the STAR mesh by closest-point projection as before.
%
%
%
We use the corners registered to the STAR model's rest pose as the initial rest pose shape $\restpose{V}^0$, and use barycentric interpolation to generate the initial skinning weights $\mathbf{W}^0$. 
Note that the number of vertices and mesh connectivity of our body model is different from the STAR model's mesh. We use each corner on the suit as a vertex of our model, and the rest pose vertex $\restpose{v}_i$ corresponds to corner $i$ in our suit. The meshing of our body model is discussed below. We use the STAR model's joints as the initial joint location $\mathbf{J}^0$. We removed the joints that controls head, neck, toes and palm from the STAR model, resulting in $M=16$ joints. We call this model our \textit{initial body model}.

\paragraph{Model refinement} After the initialization, we further optimize the shape parameters
to obtain our refined body model that more accurately fits a specific actor. Specifically, we optimize
the skinning weights $\mathbf{W}$, the joint locations $\mathbf{J}$ and the rest pose vertex positions $\restpose{V}$. 
%
Unlike *SMPL or STAR, we do not use pose-corrective blend shapes and instead correct the shape by interpolating non-articulated displacements, discussed in Section \ref{subsec:PointCloudCompletion}. 

If $P^k, k=1,2,\dots,K$ is the set of 3D points that were reconstructed from frame $k$ and $K$ is the number of frames in the training set, we refine the body model by minimizing:
\begin{equation}
\mathcal{L}_A(\restpose{V}, \mathbf{J}, \mathbf{W}, \Theta) = \mathcal{L}_f(\restpose{V}, \mathbf{J}, \mathbf{W}, \Theta) + \lambda_g \mathcal{L}_g(\mathbf{W}) + \lambda_J \mathcal{L}_J(\mathbf{J})
\label{eq:actorTuningLoss}
\end{equation}
where $\Theta=(\theta_1, \theta_2, \dots, \theta_K)$ are the pose parameters of all the frames in the training set and $\mathcal{L}_f(\Rv{\mathbf{\tilde{V}}}, \mathbf{J}, \mathbf{W}, \Theta)$ is the fitting error term:
\ACChange{
\begin{equation}
    \mathcal{L}_f(\restpose{V}, \mathbf{J}, \mathbf{W}, \Theta) =\frac{1}{ \sum_{k=1}^K|P^k|} \sum_{k=1}^K  \sum_{\mathbf{p}_i^k \in P^k} \| D_i(\restpose{v}_i, \mathbf{J}, \mathbf{W}, \theta_k) - \mathbf{p}_i^k \|_2^2
\label{eq:fittingError}
\end{equation}
}
$\mathcal{L}_J(\mathbf{J})$ is an $l_2$ loss penalizing joint locations moving too far away from their initial positions
and $\mathcal{L}_g(\mathbf{}{W})$ is a regularization term encouraging sparsity of the skinning weights:
\begin{equation}
\mathcal{L}_g(\mathbf{W}) = \sum_{i=1 }^{N} \sum_{j=1 }^{M} g_{i,j} w_{i,j}^2
\end{equation}
where  $g_{i,j}$ is the geodesic distance from corner $i$ to the closest vertex that has non-zero initial weight for joint $\mathbf{j}_j$ in the STAR model.
The regularization weights were empirically set to \ACChange{$\lambda_g = 1000$} and \ACChange{$\lambda_J=1$} when our spatial units are millimeters.

We optimize $\mathcal{L}_A$ with an alternating optimization scheme. Starting with the initial LBS model, we first calculate pose parameters $\theta_k$ for each frame. Then we optimize $\mathbf{W}$, $\mathbf{J}$, and $\restpose{V}$ one by one, while keeping the other parameters fixed. We iterate this procedure until the error decrease becomes negligible; in our results we needed between 50-100 iterations.

\begin{figure}
    \setlength\belowcaptionskip{-3ex}
    \centering
    \includegraphics[width=0.9\columnwidth]{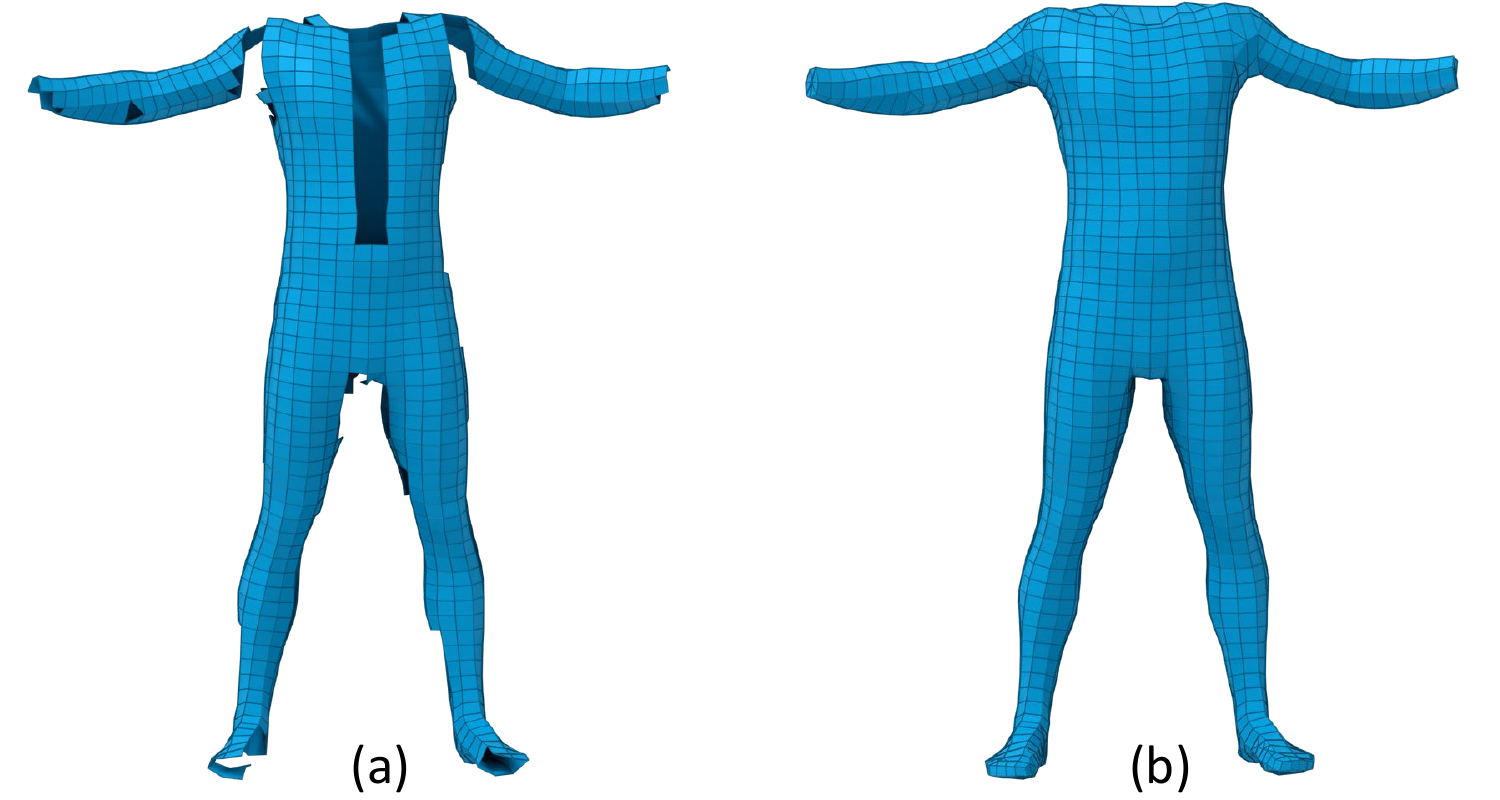}
    \caption{(a) The quad structure corresponding to our suit has holes due to the zipper and the seams; (b) The completed rest pose mesh.
    }
    \label{fig:RestposeMesh}
\end{figure}

After the optimization is finished, we mesh the rest pose vertices $\restpose{V}$. From the unique ID of each corner, we know how they were connected into quads in the suit. We manually add vertices to close the holes which come from areas of the suit such as the zipper and the seams (see \Fig{fig:RestposeMesh}a). The result is a quad-dominant mesh (\Fig{fig:RestposeMesh}b). 
\vspace{-3mm}
\subsection{Point Cloud Completion}
\label{subsec:PointCloudCompletion}
\begin{figure}
    \setlength\belowcaptionskip{-3ex}
    \centering
    \includegraphics[width=0.9\columnwidth]{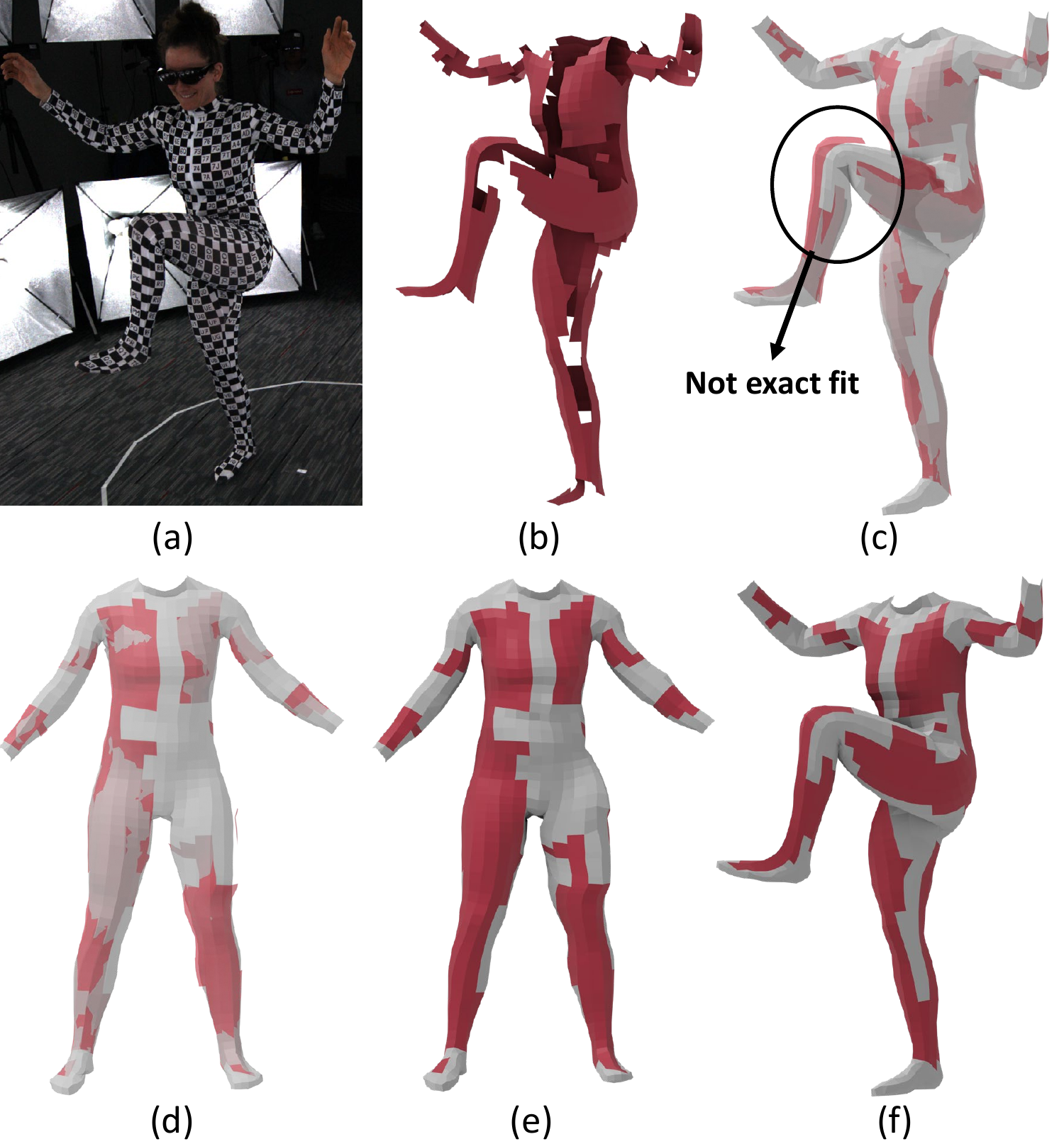}
    \caption{Our hole-filling pipeline: (a) an input image; (b) \ACChange{3D points} $P$ reconstructed from input images and connected with quads; 
    (c) our refined body model (gray) \ACChange{does not fit $P$ exactly (transparent rendering shows discrepancies)}; 
    (d) inverse skinning, mapping $P$ to the rest pose; (e) rest pose mesh interpolation, matching the inverse-skinned $P$ exactly; (f) the final result obtained by forward skinning of the interpolated rest pose mesh.
    }
    \label{fig:TrackingPointsCompletion}
\end{figure}
After the optimization, the fitting error 
(Eq.~\ref{eq:fittingError}) will drop from 13.5mm to 7.1mm on the test set; further results are reported in Section \ref{subsec:EValuateActorTuning}.
The refined LBS body model is good for representing articulated skeletal motion of the actor's body, but it does not represent well effects such as breathing or flesh deformation.
However, the non-articulated component of the motion that cannot be represented by LBS is relatively small.
Therefore, we start by applying inverse skinning transformations (also known as ``unposing'') to our observed 3D reconstructed points $\mathbf{p_i}$, see Fig.~\ref{fig:TrackingPointsCompletion}d. We denote the inverse skinning of point $i$ at pose $k$ as $D_i^{-1}(\mathbf{p}_i, \mathbf{J}, \mathbf{W}, \theta_k)$. As can be seen in Fig.~\ref{fig:TrackingPointsCompletion}d, the
$D_i^{-1}(\mathbf{p}_i, \mathbf{J}, \mathbf{W}, \theta_k)$ will not exactly match $\restpose{v}_i$ due to the non-articulated residuals.
Formally, the non-articulated displacements $\Delta\restpose{v}^k_i$ are defined as:
\begin{equation}
\mathbf{v}_i^k = D(\restpose{v}_i + \Delta\restpose{v}_i^k, \mathbf{J},  \mathbf{W}, \theta_k)
\end{equation}
The key problem of our inpainting consists in interpolating the values of $\Delta\restpose{v}_i^k$ from the observed points to the unobserved ones, in other words, predicting the unobserved non-articulated displacements, see Fig. \ref{fig:TrackingPointsCompletion}e. The modified rest pose is then mapped back by (forward) skinning to produce the final mesh, see \ACChange{Fig.} \ref{fig:TrackingPointsCompletion}f.

Our method for predicting the unobserved non-articulated displacements in the rest pose is based on the assumption of spatio-temporal smoothness.
We stack all of the rest pose displacements into a $KN \times 3$ matrix $\mathbf{X}$, where $K$ is the number of frames and $N$ the number of vertices (all vertices, both unobserved and observed ones).
%
%
We find $\mathbf{X}$ by solving the following constrained optimization problem:
\begin{equation}
\label{eq:LaplacianInterpo}
\begin{aligned}
& \min  \mathcal{L}_{\mathrm{spat}}(\mathbf{X}) + w_T \mathcal{L}_\mathrm{temp}(\mathbf{X})\\
& \text{ s.t. } \mathbf{C} \mathbf{X} = \mathbf{D}
\end{aligned}
\end{equation}
where \ACChange{$ \mathcal{L}_{\mathrm{spat}}$} is a spatial Laplacian term that penalizes non-smooth deformations of the mesh and \ACChange{$\mathcal{L}_{\mathrm{temp}}$} is a temporal Laplacian term that penalizes non-smooth trajectories of the vertices. Both of the terms are positive semi-definite quadratic forms. The parameter $w_T$ is a weight balancing these two terms which we empirically set to 100. The sparse selector matrix $\mathbf{C}$ represents the observed points (constraints) and $\mathbf{D}$ their unposed 3D positions for each frame (each frame may have a different set of observed points).
\ACChange{
Specifically, we define $\mathcal{L}_{\mathrm{spat}}$ as:
    \begin{equation}
    \mathcal{L}_{\mathrm{spat}} = \sum_{i=1}^{3} \mathbf{X}_i^T \mathbf{L} \mathbf{X}_i 
    \end{equation}
    where $\mathbf{L}$ is cotangent-weighted Laplacian of the rest pose and $\mathbf{X}_i$ is the $i$-th column of $\mathbf{X}$. We found this quadratic deformation energy to be sufficient because our non-articulated displacements in the rest pose are small, though in future work it would be possible to explore non-linear thin shell deformation energies.
} 
For $\mathcal{L}_{\mathrm{temp}}$, we use 1D temporal Laplacian which corresponds to acceleration:
$\|\Delta\restpose{v}_i^{k-1} - 2 \Delta\restpose{v}_i^{k} + \Delta\restpose{v}_i^{k+1}\|_2^2$. The $\mathcal{L}_{\mathrm{spat}}$ operator is applied to all frames independently, and $\mathcal{L}_{\mathrm{temp}}$ is applied to all vertices independently. However, their weighted combination in Eq.~\ref{eq:LaplacianInterpo} introduces spatio-temporal coupling, allowing one observed point to affect unobserved points through both space and time.

The optimization problem Eq. \ref{eq:LaplacianInterpo} is a convex quadratic problem subject to equality constraints, which we transform to a linear system (KKT matrix) and solve. The only complication is that when processing too many frames, the KKT system can become too large. For example, with $K = 5000$ frames, the KKT matrix becomes approximately $10^7\times 10^7$. Even though the KKT matrix is sparse, the linear solve becomes costly. To avoid this problem, we observe that smoothing over too many frames is not necessary and introduce a windowing scheme, decomposing longer sequence into 150-frame windows and solve them independently. To avoid any non-smoothness when transitioning from one window to another, the \ACChange{150-frame windows} overlap by 50 frames. After solving the problem in Eq.~\ref{eq:LaplacianInterpo} for each window separately, we smoothly blend the overlapping 50 frames to ensure smoothness when transitioning from one window to the next one.\looseness=-1

In this section we considered only off-line hole filling, where we can infer information from future frames. This approach would not be applicable to real-time settings where future frames are not available.

\section{Results}
\label{sec:Results}
\small
\begin{table}[ht]
\caption{Composition of our training data. The original training set is the training set before data augmentation from 20 annotated images. The test set A is from 4 annotated images of our 3 actors wearing the hand-drawn suit who appeared in the training set. The test set B contains images of fourth actor wearing a suit with printed font.} 
\centering 
\begin{tabular}{l l l l} 
\hline 
Type of Data & \Cornerdet{} & \Rejector{} & \RecogNet{} \\ [0.5ex] 
\hline 
Original Training Set & 667320 & 21257 & 7402 \\ 
Augmented  Training Set & 5678934 & 121060 & 118432 \\
Synthetic Training Set & \textit{NA} & \textit{NA} & 214471 \\
Test set A & 136500 & 3372 & 1245 \\
Test set B (printed suit)  & 134641 & 3523 & 900 \\[1ex]
\hline 
\end{tabular}
\label{table:DataComposisiton} 
\end{table}
\normalsize

We first discuss the details of our CNN training process and the results. To prepare the training data, we manually annotated 24 randomly selected images ($4000\times 2160$) of our three actors wearing the handwritten suits. 
Out of the 24, we withheld 4 images as test set A. To evaluate our system's ability to generalize to a different suit and person, we also captured a short sequence of a fourth actor wearing the printed suit and annotated 4 images from this sequence as test set B. Note that our trained process has never seen the font in the printed suit.
Table \ref{table:DataComposisiton} shows the total numbers of images used for training our CNNs. 
As shown in the first row of Table \ref{table:DataComposisiton}, the original training set (without data augmentation) for \RecogNet{} is much smaller compared to \Cornerdet{} and \Rejector{}, because of the limited number of valid quads in each of our annotated images. To improve the classification performance of \RecogNet{}, we used synthetically generated images to complement the real data, as discussed in \AppendixSynthData{}. The synthetic data contain 214471 crops ($104\times 104$), which significantly improved the robustness of \RecogNet{}, see Section \ref{subsec:EvaluateCNNS}.


We train our CNNs using Tensorflow \cite{tensorflow2015-whitepaper} using a single NVIDIA Titan RTX; for each of our CNNs, an overnight run is typically enough to converge to good results using the Adam optimizer. 
%
After our CNNs have been trained, we run inference on a PC with an i7-9700K CPU and \ACChange{an} NVIDIA GTX 1080 GPU. With a $4000\times 2160$ input image, an inference pass of \Cornerdet{} takes ~300ms, generating candidate quads ~10ms, \Rejector{} takes 1-2s to classify all of the candidate quads ($104 \times 104$) and \RecogNet{} takes ~5ms to recognize the valid quads. The computational bottleneck is the \Rejector{} due to the large number of candidate quads; this could be improved in the future by a more aggressive culling of candidate quads.
For each frame, the time for 3D reconstruction is negligible, taking less than 1ms for all points. Even though we used only one computer and processed our image sequences off-line, we would like to point out that our method for extracting 3D labeled points from multi-view images is embarrassingly parallel, because each frame \ACChange{and even each input crop for our CNNs} can be processed independently. Coupling through time is introduced only in the final hole-filling step (Section \ref{subsec:PointCloudCompletion}). The time for solving the sparse linear system (Eq. \ref{eq:LaplacianInterpo}) for a 150 frames window is about 10s.

\ACChange{We captured motion sequences of three actors, one male and two females. One of the female actor wears the small suit and the other two actors wear the medium suit. For each actor, we captured about 12,000 frames (at 30FPS) of raw image data consisting of 1) camera calibration, 2) 6000 frames of calisthenics-type sequence intended for body model refinement (also serving as a warm-up for the actor), 3) the main performance. Each frame consists of 16 images from our multicamera setup. It took about 300 hours to process all of the 576,000 images (4.6 TB) using one computer.} 
\subsection{Evaluation of CNNs}
\label{subsec:EvaluateCNNS}
\begin{figure}     
    \setlength\belowcaptionskip{-1ex}
    \centering
    \includegraphics[width=0.9\columnwidth]{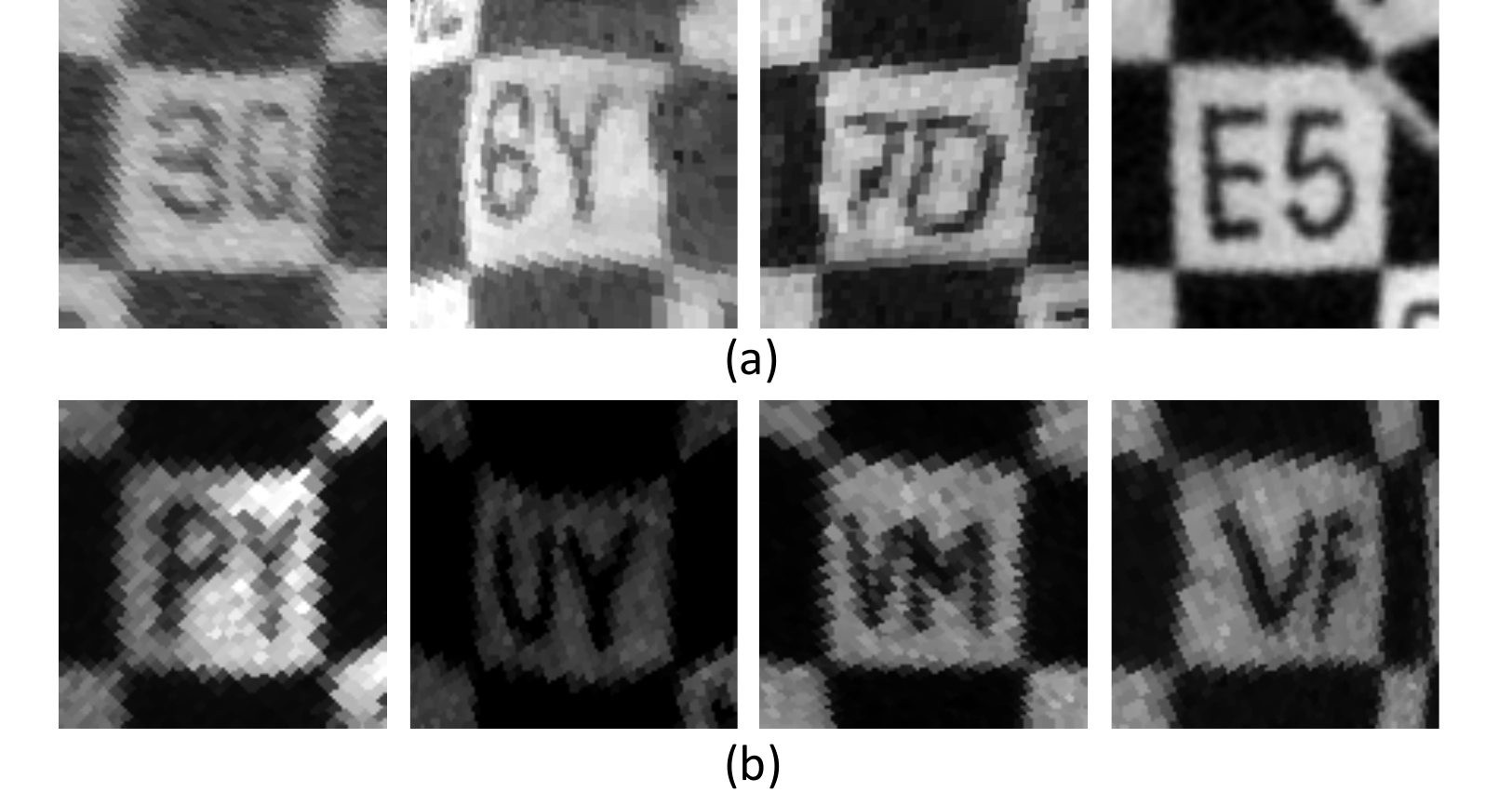}
    \caption{Examples of errors made by \Rejector{}. (a) False positives; even though the codes are legible, these samples were labeled negative due to slight image imperfections. (b) False negatives, labeled positive but close to the decision boundary.}
    \label{fig:RejectorNet_FalsePos_FalseNeg}
\end{figure}

\paragraph{\Cornerdet{}} There are two parts of the \Cornerdet{}'s output: 1) classification response that predicts whether there is a valid corner in the \ACChange{center $8\times 8$ window of the} input $20\times 20$ image and 2) its subpixel coordinates (or arbitrary values of a corner is not present).
In Table \ref{table:confuMatCornerdet} we summarize the results for both classification and localization errors. The localization error is measured by the distance in pixels between the predicted corner location and the manually annotated corner location.
The overall classification accuracy for \Cornerdet{} is 99.393\% on the training set and \ACChange{99.510}\% on the test set A. The fact that \Cornerdet{} works better on \ACChange{the} test set A supports our hypothesis that more aggressive data augmentation results in worse performance on the training set but better performance on the test set. On the test set A, the average localization error is 0.21 pixels and 99\% of the corner localizations achieve error 0.6361 pixels or less, which is remarkably low. Similar accuracy was also achieved on the test set B, which is from the printed suit, which means \Cornerdet{} can generalize to a suit that it has not been trained on. With our camera setup, 1 pixel error corresponds to approximately 1mm of 3D error for an actor 2 meters away from the camera. In practice, this means that our 3D reconstructed points are highly accurate, allowing us to capture minute motions such as muscle twitches or flesh jiggling.
\small
\begin{table}[htb]
\caption{Results for \Cornerdet{} on training and test sets.}

\begin{subtable}{\columnwidth}
\centering
\begin{tabular*}{0.9\linewidth}{@{\extracolsep{\fill}}c c c }
        n=5678934 & Actual True & Actual False  \\
        \hline
        Prediction True & 2.533\% &  0.587\% \\
        Prediction False & 0.02\% &  96.86\% \\
        \hline
        \end{tabular*}
        \caption{Confusion matrix on training set}
\label{subtable:confuMatCornerdetTrain}
\end{subtable}

\begin{subtable}{\columnwidth}
 \centering
\begin{tabular*}{0.9\linewidth}{@{\extracolsep{\fill}}c c c}
        n=136500 & Actual True & Actual False  \\
        \hline
        Prediction True & 1.7\% &  0.44\% \\
        Prediction False & 0.051\% &  97.81\% \\
        \hline
        \end{tabular*}
        \caption{Confusion matrix on test set A}
    \label{subtable:confuMatCornerdetTest}
\end{subtable}

\begin{subtable}{\columnwidth}
 \centering
\begin{tabular*}{0.9\linewidth}{@{\extracolsep{\fill}}c c c}
        n=134641 & Actual True & Actual False  \\
        \hline
        Prediction True & 1.214\% &  0.34\% \\
        Prediction False & 0.021\% &  98.432\% \\
        \hline
        \end{tabular*}
        \caption{Confusion matrix on test set B (printed suit)}
    \label{subtable:confuMatCornerdetTest2}
\end{subtable}

\begin{subtable}{\columnwidth}
 \centering
\begin{tabular*}{0.9\linewidth}{@{\extracolsep{\fill}}c c c c}
     Data set  & Max & Mean & Median  \\
        \hline
        Training set& 1.959 &  0.1793 &  0.1448  \\
        Test set A & 0.9485 & 0.2121 &  0.1904 \\
        Test set B (printed suit)& 1.081 & 0.2321 &  0.1866 \\
        \hline
        \end{tabular*}
        \caption{Max/mean/median of corner localization error in pixels}
    \label{subtable:confuMatCornerdetLocalization}
\end{subtable}
 \centering
 
\begin{subtable}{\columnwidth}
 \centering
\begin{tabular*}{0.9\linewidth}{@{\extracolsep{\fill}}c c c c c}
     Data set  & 95\% & 99\%  & 99.9\% & 99.99\%   \\
        \hline
        Training set& 0.4613 &  0.6611 &  0.9115 & 1.166  \\
        Test set A& 0.5151 & 0.6361 &  0.8912 & 0.9428 \\
        Test set B (printed suit)& 0.5831 & 0.6331 &  0.9684 &  1.083 \\
        \hline
        \end{tabular*}
        \caption{Percentiles of corner localization error in pixels}
    \label{subtable:confuMatCornerdetLocalizationPercentiles}
\end{subtable}
\label{table:confuMatCornerdet}
\end{table}
\normalsize
\paragraph{\ACChange{\Rejector{}}} The confusion matrices of a trained \Rejector{} network are reported in Table \ref{table:confuMatRejector}. The overall classification accuracy for \Rejector{} is 99.723\% on the training set, 99.704\% on the test set A, and 99.31\% on the test set B. 
From the confusion matrix, we can observe that we have more false positives than false negatives. 
The reason is that we intentionally annotated the training data conservatively. As shown in Fig. \ref{fig:RejectorNet_FalsePos_FalseNeg}a, quads with even slight imperfections were labeled as negative examples. 
This results in \RecogNet{} reporting more false positives, but the \RecogNet{} actually inherits the conservative nature of the annotations; in practice, \RecogNet{} only rarely accepts a low-quality quad image.
This nature is well demonstrated by its performance on test set B. Because the \Rejector{} have never seen the font used on the printed suit, due to its conservative nature it tends to reject a considerable number of valid quads (around 23\% of valid quads are rejected). On the other hand, the \Rejector{} returns no false positives, i.e., no invalid quads were falsely accepted.
\small
\begin{table}[htb]
\caption{Confusion matrix for \Rejector{}'s results on training and test set.}
\begin{subtable}{\columnwidth}

\centering
\begin{tabular*}{0.9\linewidth}{@{\extracolsep{\fill}}c c c }
        n=121060 & Actual True & Actual False  \\
        \hline
        Prediction True & 13.133\% &  0.243\% \\
        Prediction False & 0.034\% &  86.59\% \\
        \hline
        \end{tabular*}
        \caption{Confusion matrix on training set}
\label{subtable:confuMatRejectorTrain}
\end{subtable}%

\begin{subtable}{\columnwidth}
 \centering
\begin{tabular*}{0.9\linewidth}{@{\extracolsep{\fill}}c c c}
        n=3372 & Actual True & Actual False  \\
        \hline
        Prediction True & 1.305\% &  0.297\% \\
        Prediction False & 0.0\% &  98.399\% \\
        \hline
        \end{tabular*}
        \caption{Confusion matrix on the test set A}
    \label{subtable:confuMatRejectorTest}
\end{subtable}

\begin{subtable}{\columnwidth}
 \centering
\begin{tabular*}{0.9\linewidth}{@{\extracolsep{\fill}}c c c}
        n=3523 & Actual True & Actual False  \\
        \hline
        Prediction True & 2.286\% &  0\% \\
        Prediction False & 0.6812\% &  97.03\% \\
        \hline
        \end{tabular*}
        \caption{Confusion matrix on the test set B (printed suit))}
    \label{subtable:confuMatRejectorTest2}
\end{subtable}
\label{table:confuMatRejector}
\end{table}
\normalsize
\paragraph{\RecogNet{}} 
We compare the \RecogNet{} trained with/without the synthetic training set in Table \ref{table:resultsRecognizer}. Without using the synthetic training set, the \RecogNet{} had prediction accuracy of 99.522\% on the test set A. This accuracy was low and it was the main source of errors in our pipeline.
Enhanced with the synthetic training set, the prediction accuracy on the test set A increased to 99.919\% and significantly improved our results. Moreover, without using the synthetic training set, the \RecogNet{} cannot generalize well to the test set B: the prediction accuracy on it is only 85.16\%. That is because \RecogNet{} has never seen this new font on the printed suit. However, by enhancing training set with synthetic data which uses various fonts, the generalizability of \RecogNet{} has improved significantly, achieving an accuracy of 100\% on the test set B.
\small
\begin{table}[thb]
\caption{Classification accuracy for \RecogNet{} trained with/without synthetic training data.}
\begin{tabular*}{0.95\columnwidth}{lllll}
\hline
\multirow{3}{*}{\begin{tabular}[c]{@{}l@{}}With synthetic\\ training set\end{tabular}} & \multicolumn{4}{c}{Classification Accuracy}                            \\ \cline{2-5} 
                                                                                       & Real     & Synthetic & \multirow{2}{*}{Test A} & \multirow{2}{*}{Test B} \\
                                                                                       & Training & Training  &                       &                         \\ \hline
No                                                                                     & 99.940\% & NA        & 99.522\%              & 85.16\%                \\
Yes                                                                                    & 99.967\% & 94.849\%  & 99.919\%              & 100\%      \\ \hline           
\end{tabular*}
\label{table:resultsRecognizer}
\end{table}
\normalsize
\paragraph{Overall performance.} In the previous sections we reported the results of each individual CNN. To evaluate our complete corner localization and labeling pipeline (Fig. \ref{fig:CNNPipeline}), we use our 2 test sets of 8 manually annotated images ($4000 \times 2160$) where we know the ground truth positions and labels of all corners.
The images in the test set A and test set B contain 1702 and 1296 manually labeled corners, respectively.
Our 2D pipeline detected 92\% (1566 of 1702) of the ground truth corners from the test set A and 69\% (896 of 1296) of the ground truth corners from the test set B. The discarded corners corresponded to low quality quads or quads with printed fonts that are significantly different from the hand-drawn ones, which were rejected by \Rejector{}. Note that we intentionally trained the \Rejector{} to be conservative, i.e., to reject all borderline cases. Missing observations on the test set A do not represent a big problem because they can be fixed by inpainting \ACChange{(Section \ref{subsec:PointCloudCompletion})}; also, we observed that low-quality quads are often associated with inaccurate corner localization, increasing the noise in the 3D reconstruction. 
The missing observations on the test set B could be improved by enhancing the training of \Rejector{} with a few more annotated images from the new suit, or training \Rejector{} with synthetic data.
The mean corner localization error in our is 0.4607 pixels test set A and 0.4678 on the test set B, and the maximum localization error is 1.854 on the test set A and 1.031 on the test set B. Due to our conservative rejection approach, the final CNN, \RecogNet{}, made \textit{zero} mistakes on the test sets, i.e., all of the labeled corners in the two test sets were assigned the correct label.
\subsection{Comparison to previous methods}
\paragraph{\Cornerdet} We compared our \Cornerdet{} with three alternative corner detectors: Shi-Tomasi \cite{shi1994good}, Harris \cite{harris1988combined}, and deep learning based detector ``SuperPoint'' \cite{detone2018superpoint}. For Shi-Tomasi \cite{shi1994good} and Harris \cite{harris1988combined}, we use OpenCV's \cite{bradski2008learning} implementation and the recommended parameters. We used the method cv::cornerSubPix() \cite{forstner1987fast} to refine the detected corner locations to subpixel accuracy. 
\begin{figure}     \setlength\belowcaptionskip{-3ex}
    \centering
    \includegraphics[width=0.9\columnwidth]{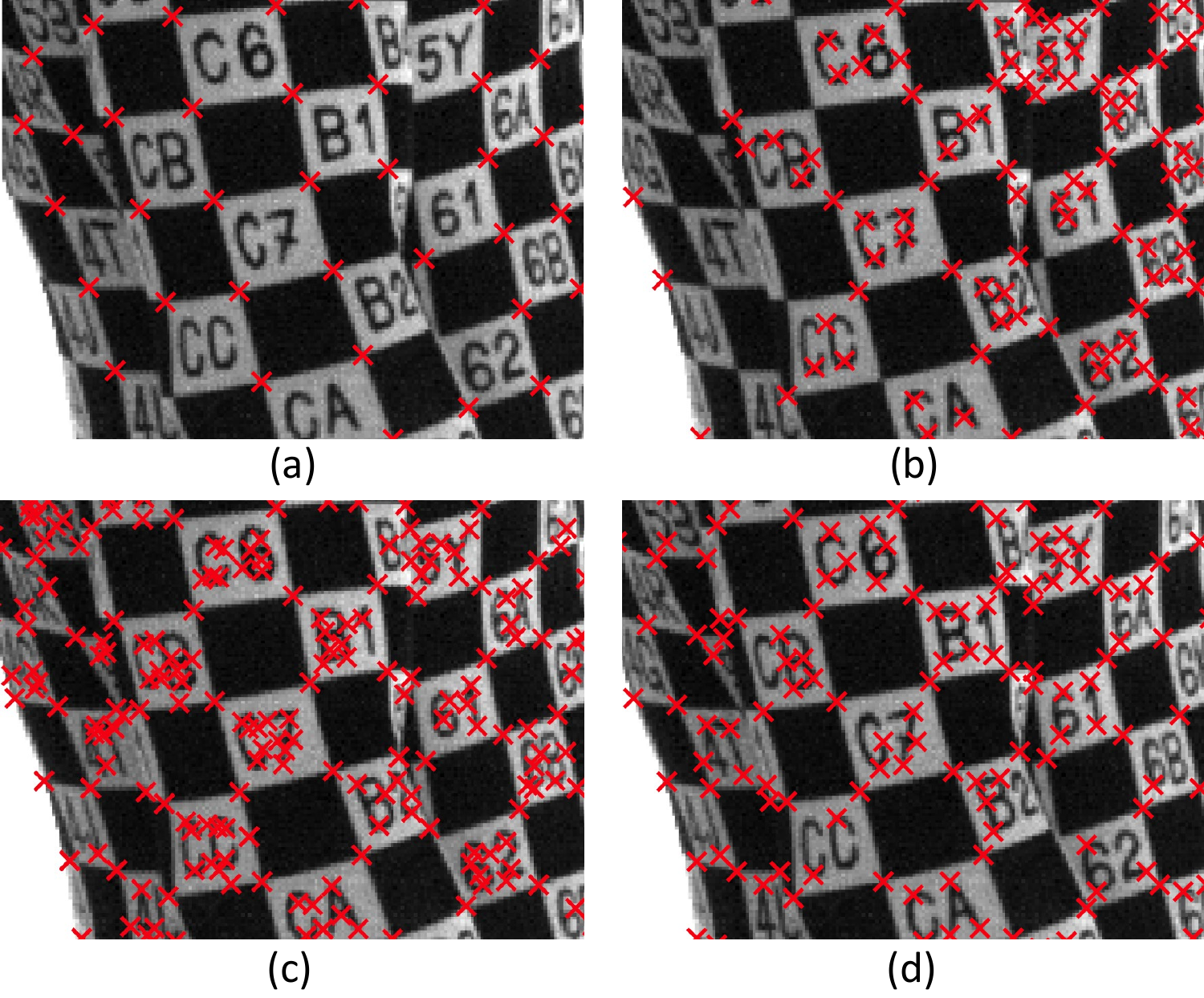}
    \caption{Comparison between different corner detectors. (a) Our \Cornerdet{}; (b) Shi-Tomasi; (c) Harris; (d) SuperPoint.
    }
    \label{fig:CornerDetectorComparison}
\end{figure}
\small
\begin{table}[]
\caption{Comparison of different corner detectors using test set B. In (a), \textit{False Negatives} means missing detection of annotated corners, \textit{False Positives} means detector yields extra \Rv{incorrect} corners that are not annotated. }
\begin{subtable}{\columnwidth}
\begin{tabular}{lllll}
                & \Cornerdet{} & Shi-Tomasi & Harris & SuperPoint \\ \hline
True Positives  & 928          & 400        & 867    & 901        \\
False Negatives & 11           & 540        & 73     & 39          \\
False Positives & 30           & 779        & 12413   & 3094        \\ \hline
\end{tabular}
\caption{Comparison of classification accuracies}
\label{subtable:cornerDetectorClassificationCmpr}
\end{subtable}
\begin{subtable}{\columnwidth}
\begin{tabular}{lllll}
                                                                      & \Cornerdet{} & Shi-Tomasi & Harris & SuperPoint \\ \hline
\begin{tabular}[c]{@{}l@{}}Localization \\ Error\end{tabular} & 0.2321    & 0.3135     & 0.3081 & 0.4937     \\ \hline 
\end{tabular}
\caption{Comparison of mean localization errors}
\label{subtable:cornerDetectorLocalizationCmpr}
\end{subtable}
\label{table:CornerDetectorComparisonTable}
\end{table}
\normalsize

A qualitative comparison of corner detection results is shown in Fig. \ref{fig:CornerDetectorComparison}.
We also \Rv{quantitatively} compare those corner detectors using the test set B \Rv{in Table \ref{table:CornerDetectorComparisonTable}}. To evaluate the classification accuracy, we match the detected corners to the annotated corners if the distance between them is less than 1.5 pixels. The classification accuracy comparison is shown (Table. \ref{subtable:cornerDetectorClassificationCmpr}). As we can see from both Fig. \ref{fig:CornerDetectorComparison} and Table \ref{subtable:cornerDetectorClassificationCmpr}, Shi-Tomasi, Harris and SuperPoints all have the same problem: they detect a lot of non-checkerboard corners (false positives), many of which are from the text on the suit. This is not surprising because those corner detectors are designed to detect general features that exists in nature, not checkerboard corners specifically. These wrong detections can slow down the processing of our pipeline because they lead to more invalid candidate quads. Also, they can confuse \Rejector{}. 
On the other hand, alternative corner detectors are missing many valid corners in comparison to \Cornerdet{}, which will results in missing observations. From Table \ref{subtable:cornerDetectorLocalizationCmpr}, we can see that those corner detectors are also inferior to \Cornerdet{} in terms of localization accuracy.\looseness=-1

\paragraph{\RecogNet{}} We compared our \RecogNet{} with Tesseract-OCR \cite{smith2007overview}. The results are shown in Table \ref{table:TextRecogCmpr}. 
Tesseract-OCR produces significantly worse results than \RecogNet{}, especially on the hand-drawn test set A. In practice, we find that Tesseract-OCR is very sensitive to stretching, distortion and noise. This is probably due to the fact that Tesseract-OCR was designed to recognize printed text with standard format rather than text on a suit worn by human which can have significant distortion due to stretching. Our dedicated \RecogNet{} performs much better in this case.
\small
\begin{table}[]
\caption{Comparison of text recognizers' prediction accuracies.}
\begin{subtable}{\columnwidth}
\centering
\begin{tabular*}{0.9\linewidth}{@{\extracolsep{\fill}}l l l }
         & \RecogNet{} & Tesseract-OCR  \\
        \hline
        test set A & 99.919\% &  19.26\% \\
        test set B & 100\% &  61.93\% \\
        \hline
        \end{tabular*}
\end{subtable}
\label{table:TextRecogCmpr}
\end{table}
\normalsize
\subsection{3D Reconstruction Accuracy}
\begin{figure}     \setlength\belowcaptionskip{-3ex}
    \centering
    \includegraphics[width=\columnwidth]{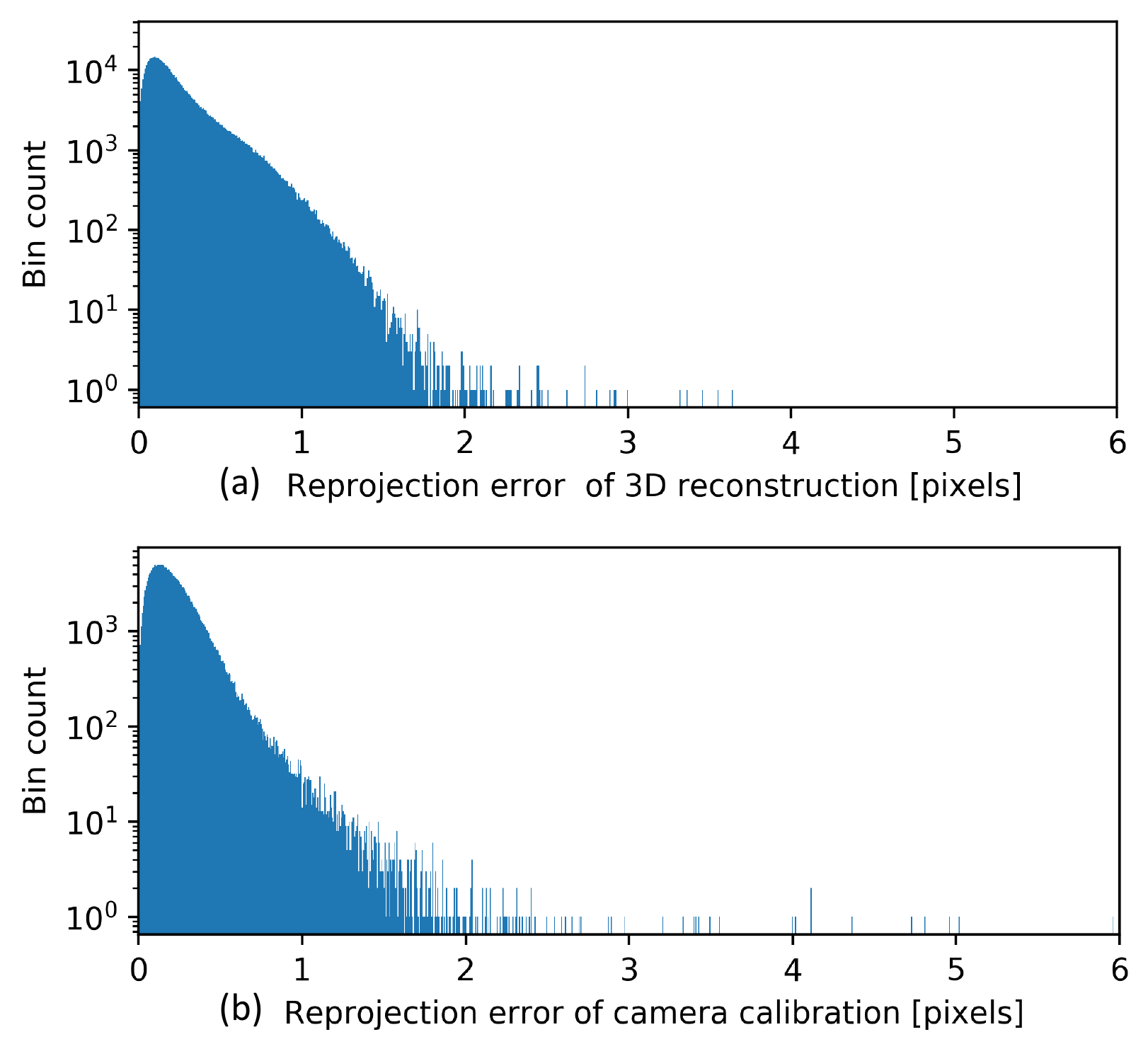}
    \caption{
    Comparison of histograms of reprojection errors in 3D reconstruction and camera calibration. 
    (a) Distribution of reprojection errors computed per camera for all the 3D reconstructed corners in 10000 consecutive frames.
    (b) The distribution of reprojection errors of camera calibration.
    We can see those two histograms look very similar.
    }
    \label{fig:ReconstructionEvaluation}
\end{figure}
\paragraph{Metrics.} Evaluating 3D reconstruction accuracy is hard, because we do not have any ground truth measurements of a moving human body. To evaluate the accuracy of our 3D reconstructed corners, we compute their reprojection errors and we compare them to the reprojection errors obtained in our camera calibration process (Section \ref{subsec:QuadProposal}). Using $f^k$ to denote the  projection function of camera $k$, if a reconstructed 3D point $\mathbf{p}_i$ is seen by camera $k$ and  $\mathbf{c}_i^k$ is the pixel location of the corresponding 2D corner $i$ in camera $k$, the reprojection error for corner $i$ in camera $k$ is defined as:
\begin{equation}
\label{eq:reprojErr}
\mathbf{e}_{i,k} =  ||  f^k(\mathbf{p}_i)-\mathbf{c}_i^k||
\end{equation}
The reprojection error for camera calibration is defined analogously, except that we use 3D calibration boards with perfect, rigid checkerboard corners and a standard OpenCV corner detector. In contrast, our corners are painted on an elastic suit worn by an actor.

\paragraph{Quantitative evaluation} We report the histograms of reprojection errors of 3D reconstruction and camera calibration in Fig. \ref{fig:ReconstructionEvaluation}. The 3D reconstruction reprojection error is computed per camera for all the reconstructed points in a consecutive sequence of 10000 frames. The calibration reprojection error was computed on 448 frames that we use to calibrate the cameras, where we wave a $9\times 12$ calibration board in front of our cameras.
In \ref{fig:ReconstructionEvaluation}, we can see the two error distributions look very similar, which means the reprojection errors of our 3D reconstruction results have similar statistics as the reprojection errors in camera calibration. We cannot expect to obtain lower reprojection errors than camera calibration. 

Table \ref{table:ReprojErrsPercentiles} shows the percentiles of all the reprojections errors in 10000 frames that we use to evaluate the 3D reconstruction. 99\% of the reprojection errors is less than 1.009 \Rv{pixels}, which is remarkably accurate given the high resolution of our images ($4000 \times 2160$).  
\begin{table}[htb]
\caption{Percentiles of reprojection errors computed per camera for all the reconstructed points in a consecutive sequence of 10000 frames.}
\small
\begin{subtable}{\columnwidth}
\centering
\begin{tabular*}{0.9\linewidth}{@{\extracolsep{\fill}}c c c c}
        \hline
      95\% & 99\%  & 99.9\% & 99.99\%   \\
        \hline
         0.6979 &  1.009 &  1.409 & 3.376  \\

        \hline
        \end{tabular*}
\end{subtable}
\label{table:ReprojErrsPercentiles}
\end{table}
\normalsize
\begin{figure}     \setlength\belowcaptionskip{-3ex}
    \centering
    \includegraphics[width=0.9\columnwidth]{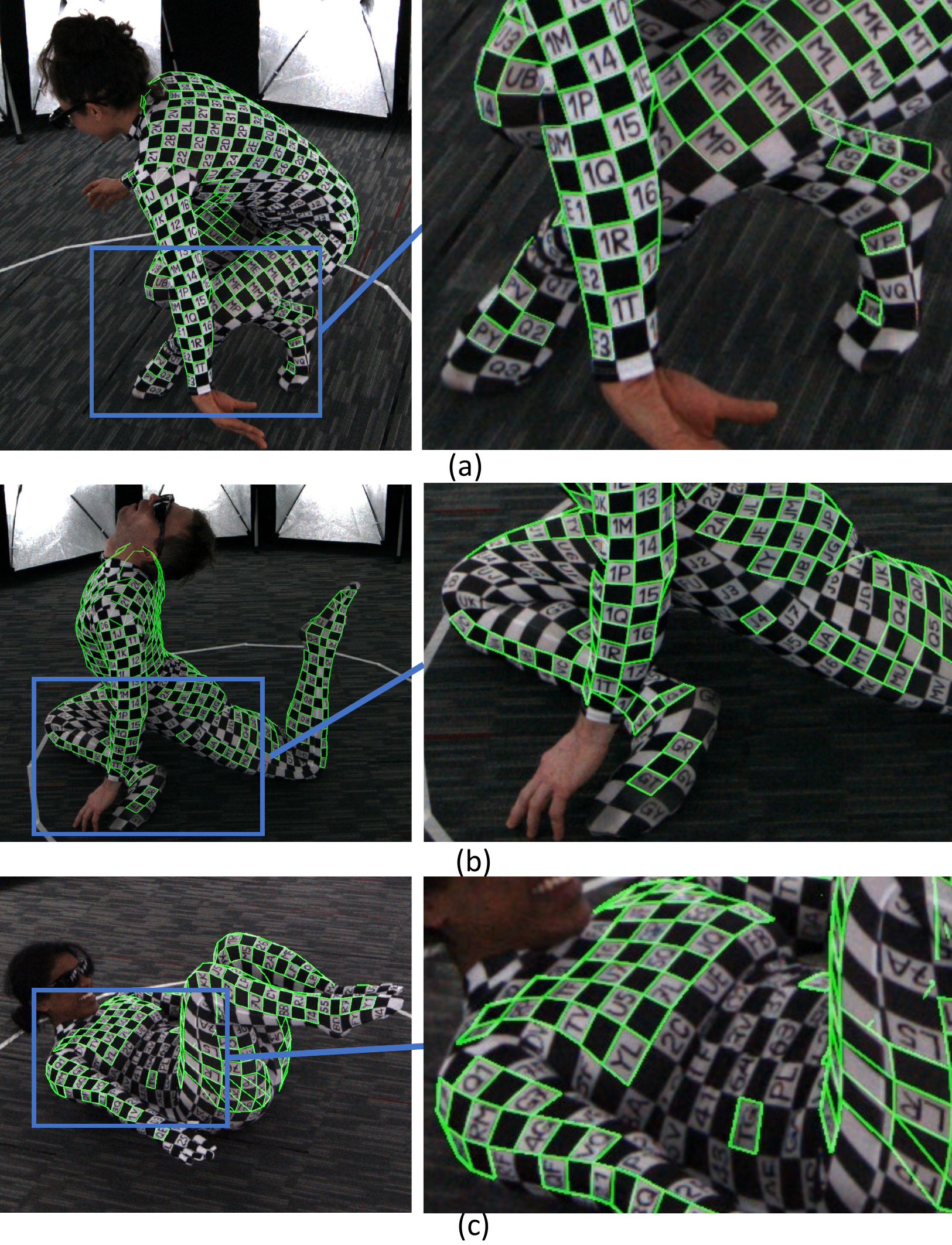}
    \caption{
        Our results in challenging poses. Our reconstructed mesh (visualized as green wireframe) closely matches the checkerboard pattern in the original input images (background). Note the successful isolated code recognitions \Rv{on} the feet in (a) and (b).
    }
    \label{fig:QualitativeResultsRecon}
\end{figure}
\paragraph{Qualitative evaluation} Fig. \ref{fig:QualitativeResultsRecon} shows challenging cases where there are significant self occlusions. We mesh the reconstructed point cloud using the rest pose mesh structure introduced in Section \ref{subsec:bodyModel} by preserving the observed faces in the rest pose mesh (see Fig.~\ref{fig:RestposeMesh}b). Then we project the reconstructed mesh back to the image using the camera parameters, which gives us the green wireframe in Fig. \ref{fig:QualitativeResultsRecon}. We can see that the mesh wireframe aligns very closely with the checkerboard pattern on the suit. Another important observation is that even despite large occlusions, our method can still obtain correctly labeled corners as long as the entire two-letter code is visible see, e.g., the foot and calf in Fig.~\ref{fig:QualitativeResultsRecon}a,b. In Fig.~\ref{fig:QualitativeResultsRecon}c, we can see that the conservative \Rejector{} correctly rejects the wrinkled quads in the belly region, since reading the codes would be difficult or impossible.
\subsection{Evaluation of Body Model Refinement}
\label{subsec:EValuateActorTuning}
\begin{figure}     \setlength\belowcaptionskip{-3ex}
    \centering
    \includegraphics[width=\columnwidth]{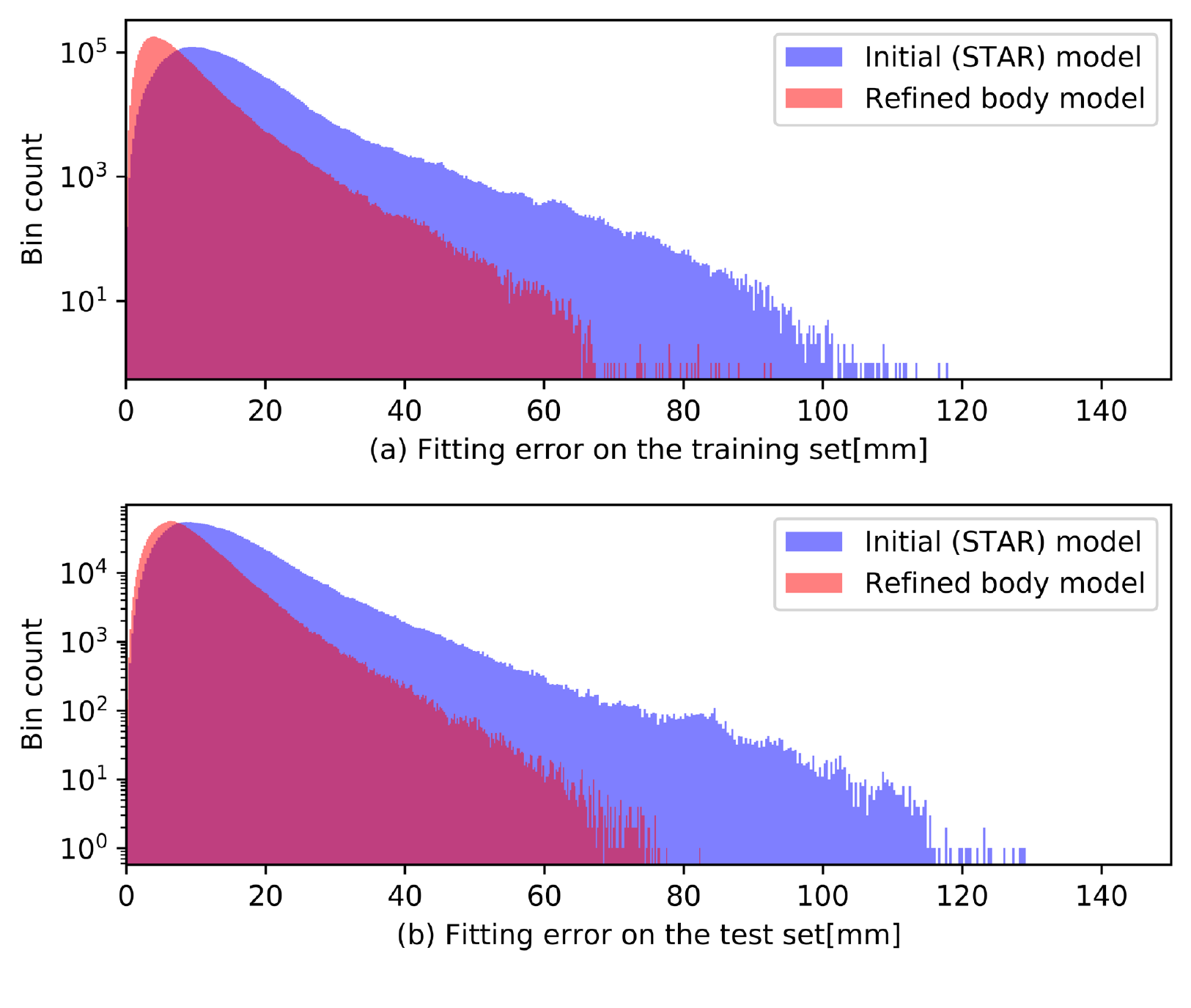}
    \caption{
       Histograms of fitting errors of the initial body model and the refined body model on the training set and the test set. 
    }
    \label{fig:ActorTuningHistograms}
\end{figure}
\begin{figure}     \setlength\belowcaptionskip{-3ex}
    \centering
    \includegraphics[width=\columnwidth]{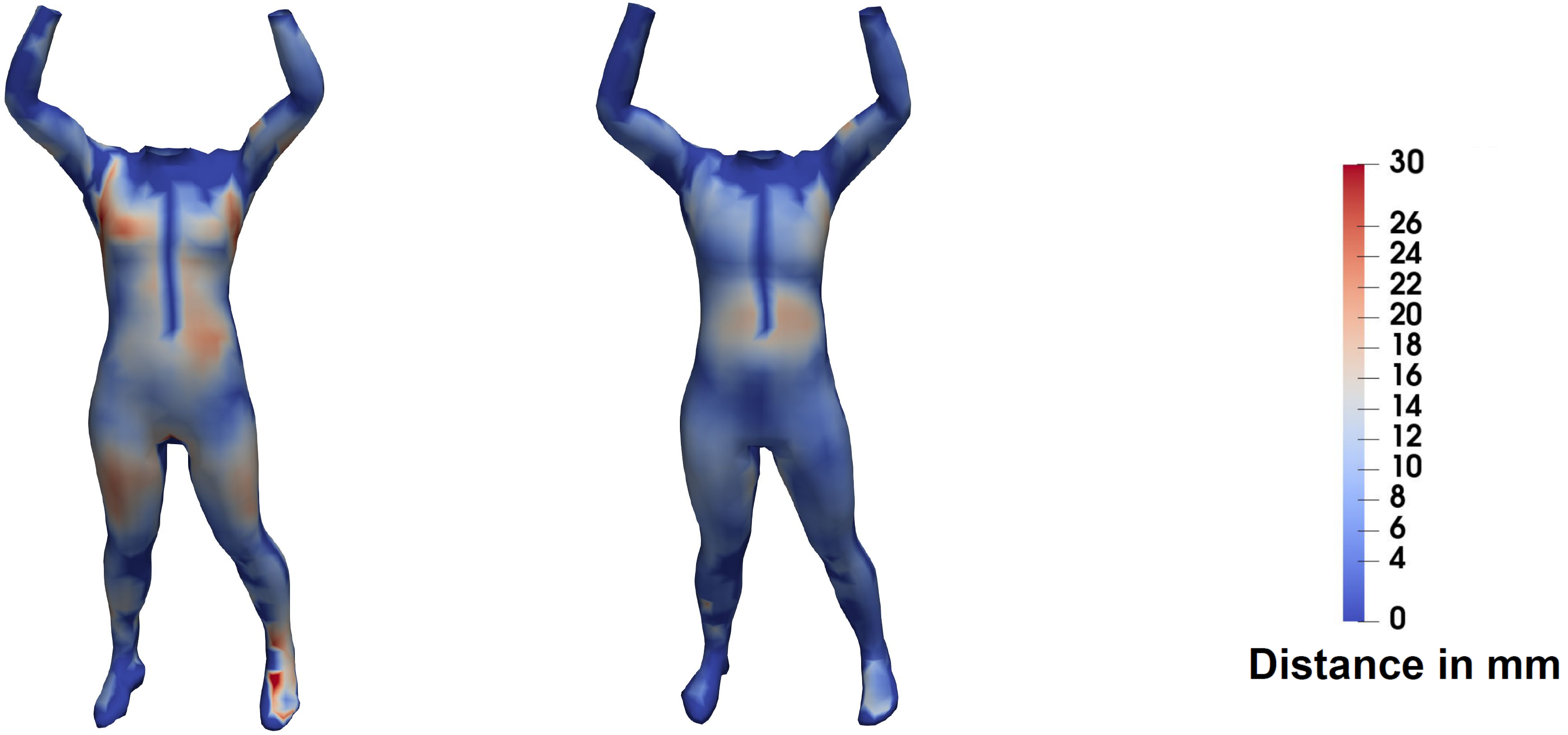}
    \caption{
      The fitting errors on the body model; left: an initial body model which is just a re-meshing of the STAR model; right: our final refined body model. 
    }
    \label{fig:ActorTuningColorMaps}
\end{figure}

To refine our actor model, we record a 6000 frames training sequence. After the body model refinement we select another 3000 frames corresponding to motions different from the ones in the training set. The fitting error is defined as the distance between the vertices of the deformed body model and the actual 3D reconstructed corners.
We compare the fitting errors between the initial model, which is just a remeshing of the STAR model (see Section \ref{subsec:bodyModel}), and the refined body model, which was optimized on the training set (see Section \ref{subsec:PointCloudCompletion}).
Fig. \ref{fig:ActorTuningHistograms} shows the distribution of fitting errors per vertex of the initial body model and the refined body model on the training set and the test set.
We can see in both data sets, the refined body model is much more accurate. Specifically, the body model refinement reduces the average fitting error from 13.6mm to 5.2mm on the training set and from 13.5mm to 7.1mm on the test set.
Fig.~\ref{fig:ActorTuningColorMaps} visualizes the fitting errors on the body model before and after body model refinement in one example frame using a heat map. 
\subsection{Evaluation with Optical Flow}
To quantify the accuracy of the 3D reconstruction of the entire body, we compare renderings of a textured mesh with the original images using optical flow \cite{bogo2014faust, bogo2017dynamic}. First, we need to create a suit-like texture for our body mesh (Fig. \ref{fig:RestposeMesh}b). 
We create a standard UV parametrization for our mesh and generate the texture from 10 hand picked frames using a differentiable renderer \cite{ravi2020pytorch3d}; though this is just one possible way to generate the texture \cite{bogo2017dynamic}.
We render the textured body mesh with back face culling enabled and overlay it over clean plates (i.e., images of the capture volume without any actor).
The virtual camera parameters are set to our calibration of the real cameras. The optical flow is computed from the synthetic images to the \ACChange{undistorted} real images using FlowNet2 \cite{ilg2017flownet} with the default settings.
\begin{figure}     \setlength\belowcaptionskip{-3ex}
    \centering
    \includegraphics[width=\columnwidth]{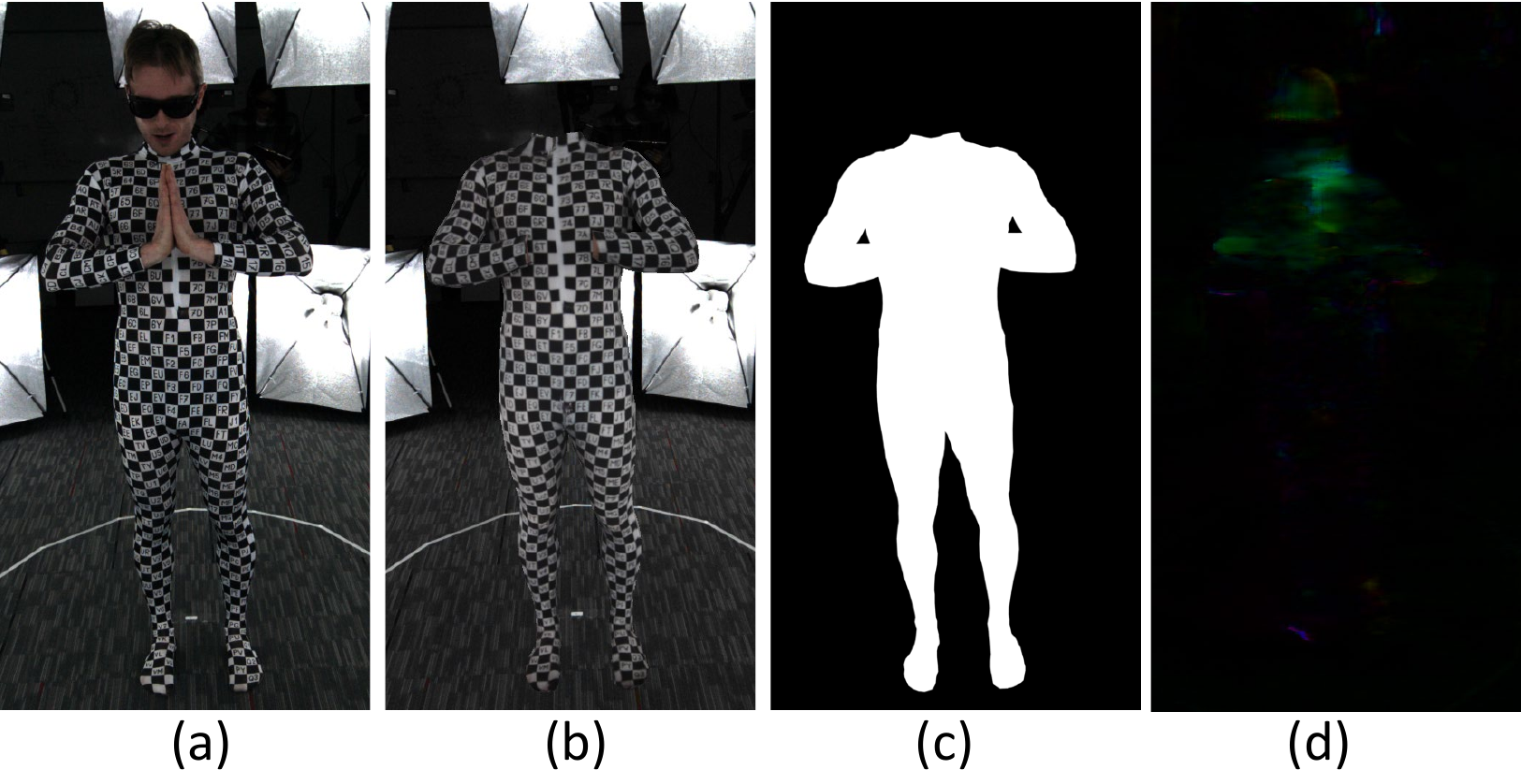}
    \caption{
        (a) The input image. (b) The synthetic image rendered from our body mesh. (c) The foreground mask of our body mesh. (d) The optical flow between the synthetic image (b) and the real one (a). The angle of flow is visualized by hue and the magnitude of flow by value in HSV color model. 
    }
    \label{fig:OpticalFlowGeneration}
\end{figure}
Because our mesh does not have the hands and the head, we first render a foreground mask of our body mesh (Fig. \ref{fig:OpticalFlowGeneration}c).
We only evaluate the optical flow on the region covered by the foreground mask to exclude the hands, the head and the background.
The foreground mask cannot exclude the hands and the head when they are occluding the body (as in Fig.~\ref{fig:OpticalFlowGeneration}a, b) but, fortunately, the optical flow is robust to missing parts (see Fig.~\ref{fig:OpticalFlowGeneration}d). 
\begin{figure}
    \setlength\belowcaptionskip{-3ex}
    \centering
    \includegraphics[width=\columnwidth]{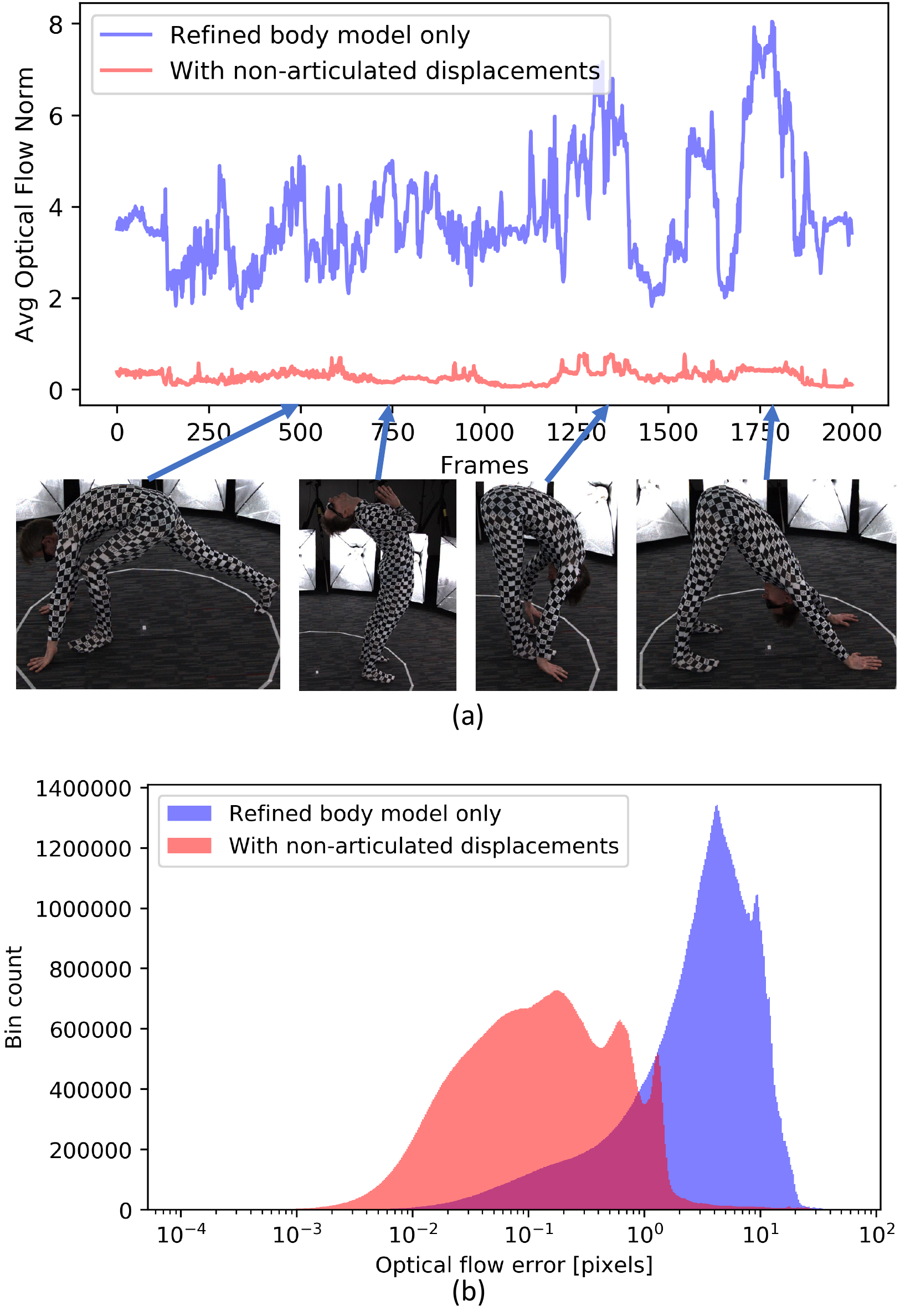}
    \caption{
        Optical flow errors. The blue curves correspond to a low-dimensional refined body model only; the red curves correspond to our final results including non-articulated displacements. (a) \ACChange{The} plot of average optical flow norm in a motion sequence of 2000 frames (we show four example frames below the graph). (b) Histograms of per-pixel optical flow norms for the same sequence.
    }
    \label{fig:OpticalFlowAnalysis}
\end{figure}
We use optical flow to compare the original images with two types of renders: 1) our low-dimensional refined body model \ACChange{(the gray mesh in Fig. \ref{fig:TrackingPointsCompletion}c, which does not fit the reconstructed corners exactly)}, 2) our final result after adding non-articulated displacements \ACChange{(Fig. \ref{fig:TrackingPointsCompletion}f}).
Fig.~\ref{fig:OpticalFlowAnalysis}a plots the average optical flow norm for each frame for consecutive 2000 frames, including various challenging poses and fast motions. We can see that the  \ACChange{result with} non-articulated displacements is much more accurate than only our low-dimensional refined body model. This is mainly due to flesh deformation which is not well explained by the refined body model, especially in more extreme poses which correspond to the spikes in the blue curve in Fig.~\ref{fig:OpticalFlowAnalysis}a. The red curve corresponds to our final result which exhibits consistently low optical flow errors.

We also plot the distribution of the optical flow norm for each pixel in the foreground mask in 
Fig.~\ref{fig:OpticalFlowAnalysis}b. With our final animated mesh, 95\% of pixels have optical flow norm less than 1.20 and 99\% of pixels have optical flow norm less than 2.46. 

\section{Limitations and Future Work}
\label{sec:Limitation}

An obvious limitation of our method is the necessity of wearing a special motion capture suit. A suit can in principle slide over the skin, but we did not observe any significant sliding in our experiments because our suits are tightly fitting. If this became a problem in the future, we could increase adhesion with internal silicone patches as in sportswear, or even apply spirit gum or medical adhesives. The suit needs to be made in various sizes and fit may be a challenge for obese people. The holy grail of full-body capture is to get rid of suits and instead rely only on skin features such as the pores, similarly to facial performance capture. We tried imaging the bare skin, but with our current camera resolution ($4000 \times 2160$) we were unable to get sufficient detail from the skin. We could obtain more detail with narrower fields of view and more cameras to cover the capture volume, but then there are issues with the depth of field and hardware budgets. Additional complications of imaging bare skin are body hair and privacy concerns; our suit certainly has its disadvantages, but mitigates these issues. A significant advantage of our suit compared to traditional motion capture suits is that we do not need to attach any markers (reflective spheres, \ACChange{See Fig. \ref{fig:SuitDesign}a}). Traditional motion capture markers can impede motion or even fall off, e.g., when the actor is rolling on the ground. An intriguing direction for future work would be to enhance our suit with additional sensors, in particular EMG, IMU or pressure sensors on the feet.

In this paper we focused on the body and ignored the motion of the face and the hands. Our actors are wearing sunglasses because our continuous passive lights are too bright; the perceived brightness could be reduced by lights which strobe in sync with camera shutters, but this would require significant investments in hardware. In future work, our method could be directly combined with modern methods that capture the motion of the face and the hands \cite{joo2018total, pavlakos2019expressive, choutas2020monocular, xiang2019monocular}.
We note that our current system captures the motion of the feet, but not the individual toes.

Our current data processing is off-line only.
In the future, we believe it should be possible to create a real-time version of our system. This would require machine vision cameras tightly integrated with dedicated GPUs or tensor processors for real-time neural network inference. Each such hardware unit could emit small amounts of data: only information about the corner locations and their labels, avoiding the high bandwidth requirements typically associated with high-resolution video streams.

Another avenue for future work involves research of different types of fiducial markers that can be printed on the suit. In fact, we made initial experiments with printing on textile and sewing our own suits, which gives us much more flexibility than handwritten two-letter codes discussed in this paper. We postponed this line of research due to the Covid19 pandemic. Our pipeline for reconstructing labeled 3D points does not make any assumptions about the human body, which means that we could apply our method also for capturing the motion of clothing or even loose textiles such as a curtain.

\vspace{-2mm}
\section{Conclusions}
We have presented a method for capturing more than 1000 uniquely labeled points on the surface of a moving human body. This technology was enabled by our new type of motion capture suit with checkerboard-type corners and two-letter codes enabling unique labeling of each corner. Our results were obtained with a multi-camera system built from \ACChange{off-the-shelf} components at a fraction of the cost of a full-body 3DMD setup, while demonstrating a wider variety of motions than the DFAUST dataset \cite{bogo2017dynamic}, including gymnastics, yoga poses and rolling on the ground. Our method for reconstructing labeled 3D points does not rely on temporal coherence, which makes it very robust to dis-occlusions and also invites parallel processing. \ACChange{We provide our code and data as supplementary materials and we will release an optimized version of our code as open source.}

\begin{acks}
We thank Marianne Kavan and Katey Blumenthal for their performances and consultation on applications in medicine; to Daniel Sykora,
Gerard Pons-Moll,
Gordon Wetzstein,
Jiawen Chen,
Ross Whitaker,
Srikumar Ramalingam and
Stepan Kment
for sharing their expertise in numerous discussions; to our annotators: Aaron Carlisle, 
Andrey Myakishev, 
Cameron James Yeomans, 
Cole Perscon, 
Dimtar Divev, 
Ejay Guo, 
Junior Rojas, 
Pranav Rajan, 
Sujay Tripathy, 
Wenxian Guo, 
Wenzheng Tao, 
Will Stout, 
Xin Yu. 
This material is based upon work supported by the National Science Foundation under Grant Numbers IIS-1764071, IIS-2008915 and IIS-2008564. Any opinions, findings, and conclusions or recommendations expressed in this material are those of the author(s) and do not necessarily reflect the views of the National Science Foundation. We also gratefully acknowledge the support of Activision, Adobe, and hardware donation from NVIDIA Corporation.
\end{acks}

\bibliographystyle{ACM-Reference-Format}
\bibliography{sample-bibliography}

\appendix

\section{CNN Structures}
\subsection{\Cornerdet{}}
\label{appendix:Cornerdet}
\begin{figure}
    \centering
    \setlength\belowcaptionskip{-3ex}
    \includegraphics[width=\columnwidth]{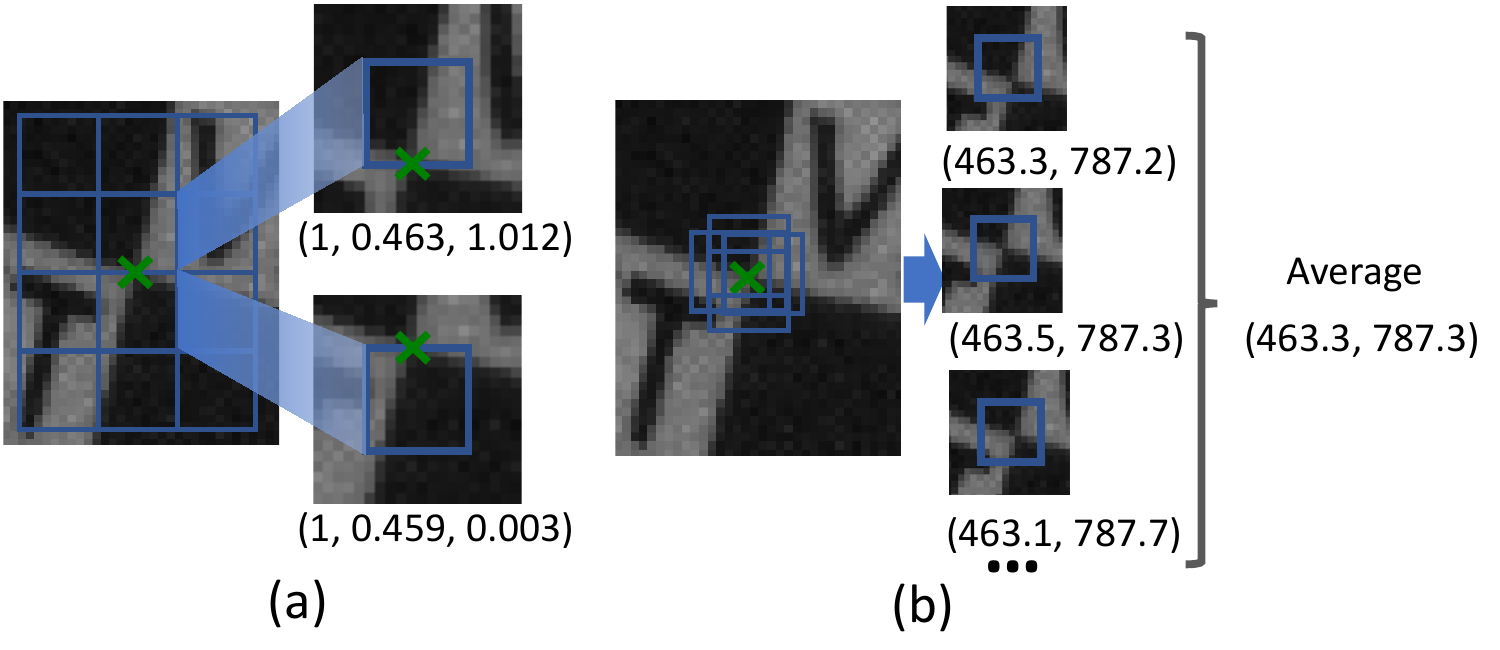}
    \caption{(a) A corner is detected twice because it is on the boundary between 2 cells. In the parentheses are the \Cornerdet{} outputs on each crop. (b) The re-crops generated for a detected corner. In the parentheses are the corner position in global pixel coordinates.}
    \label{fig:Clustering}
\end{figure}

\begin{figure*}
    \centering
    \includegraphics[width=\textwidth]{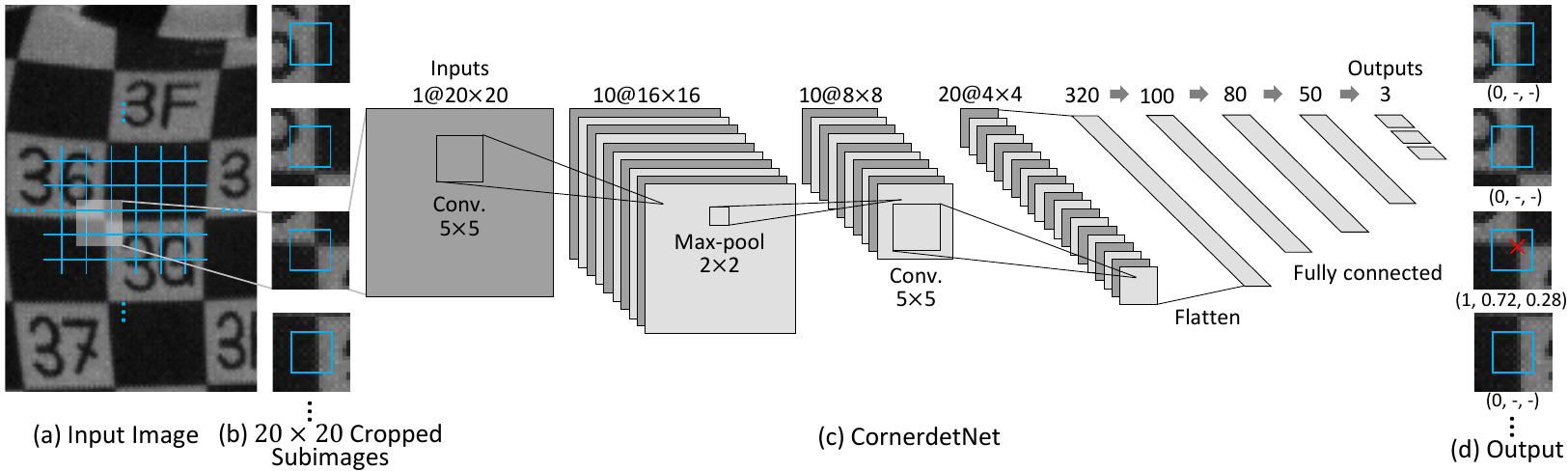}
    \caption{(a) The input image is divided into a regular grid of $8\times8$ cells. (b) Each cell is expanded by a margin and the corresponding expanded crop is passed as input to \Cornerdet{}.
    (c) The architecture of our \ACChange{\Cornerdet{}}. (d) Example outputs; even though there is a corner in the second image, it is outside of the inner $8\times8$ cell, thus the detector correctly reports 0 (no corner).
    }
    \label{fig:CornerDetector}
\end{figure*}

The input to \Cornerdet{} is the $8\times8$ cell where a corner is being sought, including a 6-pixel margins added to each side (Fig. \ref{fig:CornerDetector}b), making the input crop size $20\times20$.
These margins allow us reliably detect even the corners close to the boundaries of the $8\times8$ cell. The 6-pixel margins overlap with adjacent cells (Fig. \ref{fig:CornerDetector}b), but the $8\times8$ cells do not overlap.
The \Cornerdet{} outputs three floating-point numbers. 
The first one is a logit of a binary classifier predicting whether a corner is present or not, and the other two are normalized coordinates ($[0, 1] \times [0, 1]$) of the corner relative to the $8\times8$ cell.
The training loss for \Cornerdet{}  is:
\begin{equation}
\mathcal{L}_c(p^*, p;  \mathbf{c}^*, \mathbf{c}) =  \mathcal{L}_p(p^*, p) + \lambda_{c} p ||\mathbf{c}-\mathbf{c}^*||^2_2
\end{equation}
Where $\lambda_c$ balances the prediction loss and localization loss; we set $\lambda_c=200 $ when training \Cornerdet{}. $p^*$ represents the logit of the binary classifier, $c^*$ represents the prediction of corner location, $p$ and ${c}$ represents the ground truth respectively, and $\mathcal{L}_p(p^*, p)$ is cross entropy. 

\paragraph{Corner Clustering and Refinement}
When a corner lies exactly on the boundary of two $8\times8$ cells, it can be detected more than once, i.e., positives returned from both of the adjacent cells (Fig. \ref{fig:Clustering}a).
To fix such duplicate detections, we perform a clustering pass: if any two detected corners are too close ($< 3$ pixels), we discard the one with the lower logit value. Since this might introduce additional localization noise, we generate new crops randomly perturbed around the original corner positions, run localization on each of these crops and average the results in global pixel coordinates, see Fig. \ref{fig:Clustering}b. 
This helps especially when corners are crossing the boundaries of the $8\times8$ cells (Fig. \ref{fig:Clustering}a).

\subsection{Quad Classifiers}
\label{appendix:QuadClassifiers}
\begin{figure*}
\setlength\belowcaptionskip{-3ex}
    \centering
    \includegraphics[width=\textwidth]{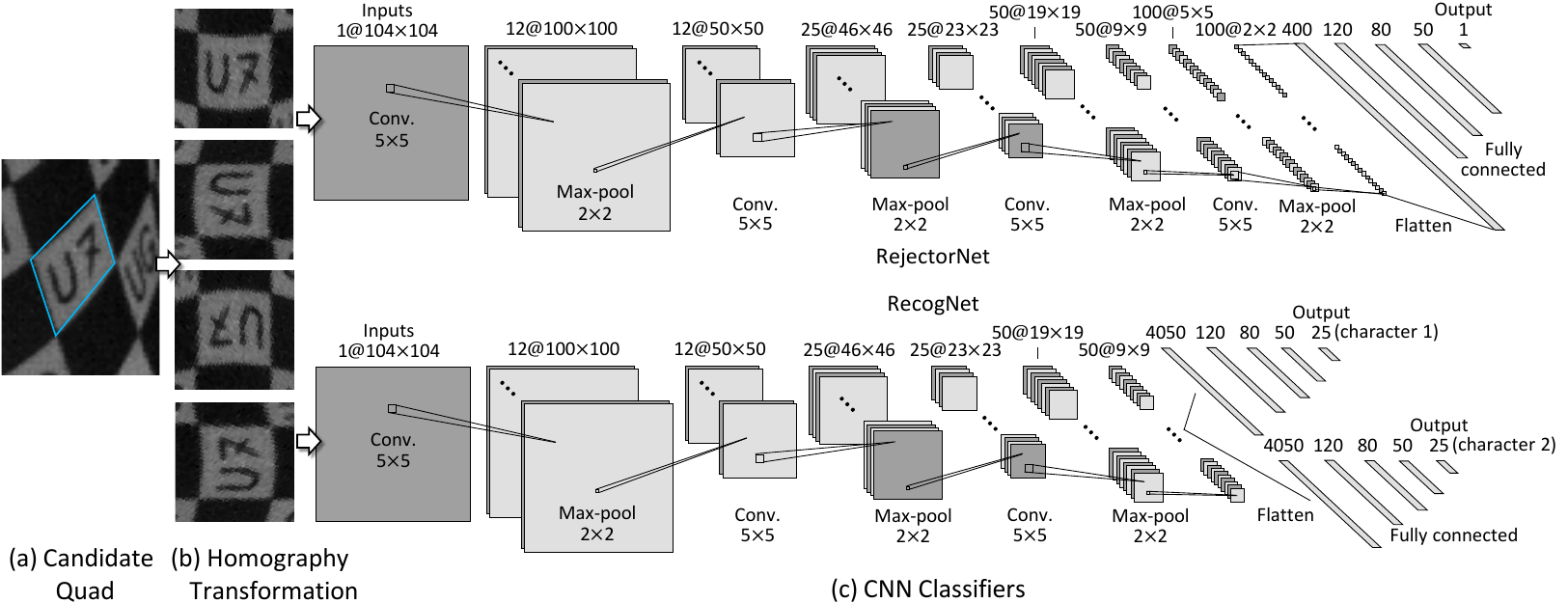}
    \caption{Our quad processing pipeline: (a) Example candidate quad.
    (b) The undistorted candidate quads with margin; these images are input to our quad classifiers. (c) The architectures of \Rejector{} and \RecogNet{} CNNs.
    }
    \label{fig:QuadClassifiers}
\end{figure*} 

The architectures of \Rejector{} and \RecogNet{} are shown in Fig.~\ref{fig:QuadClassifiers}.
\paragraph{Candidate quad generation.} It would be wasteful to enumerate all 4-tuples of corners for further processing by neural networks. Therefore, we first apply simple criteria to filter out quads that cannot contain a valid code.
We start by iterating over all the corners, and for each corner, we select three other corners within a bounding box. When connecting corners into a quad, we ensure that each quad is convex, clock-wise oriented and unique.
Additional filtering criteria include geometric criteria and image based criteria: geometric criteria constrain the area, maximum/minimum edge-lengths and maximum/minimum angles of the generated candidate quad; image based criteria constrain the average intensity, and standard deviation of all the pixels in the generated candidate quad.
To obtain the range for each criterion, we gather statistics for each of those quantities in the training dataset Section~6 and create conservative intervals to ensure that we cannot mistakenly reject any valid quad.  The candidate quads that pass all of these early rejection filters are transformed using homography and passed to further processing to quad classifier neural networks. 
\paragraph{Why separate \Rejector{} and \RecogNet{}?}  We considered combining the two networks into one, but we found that network training is easier if we treat each problem separately. Specifically, the \Rejector{} should perform quality control of a $104 \times 104$ standardized image, including rejection of errors made by \Cornerdet{} (Fig. \ref{fig:QuadClassifiersDataAug}b). Because we prefer missing observations to errors, we train \Rejector{} to be conservative and reject any inputs of dubious quality. 
The second network, \RecogNet{}, has to recognize two characters in any image. We can make \RecogNet{} more reliable by training it even on very difficult input images, enhancing the robustness of the entire pipeline. The details of our training process and data augmentation are discussed in Section~\ref{subsec:DataAug}.

\section{Data Acquisition}
\subsection{Data Conversion}
\label{appendix:DataConversion}
\begin{figure}
\setlength\belowcaptionskip{-3ex}
    \centering
    \includegraphics[width=0.9\columnwidth]{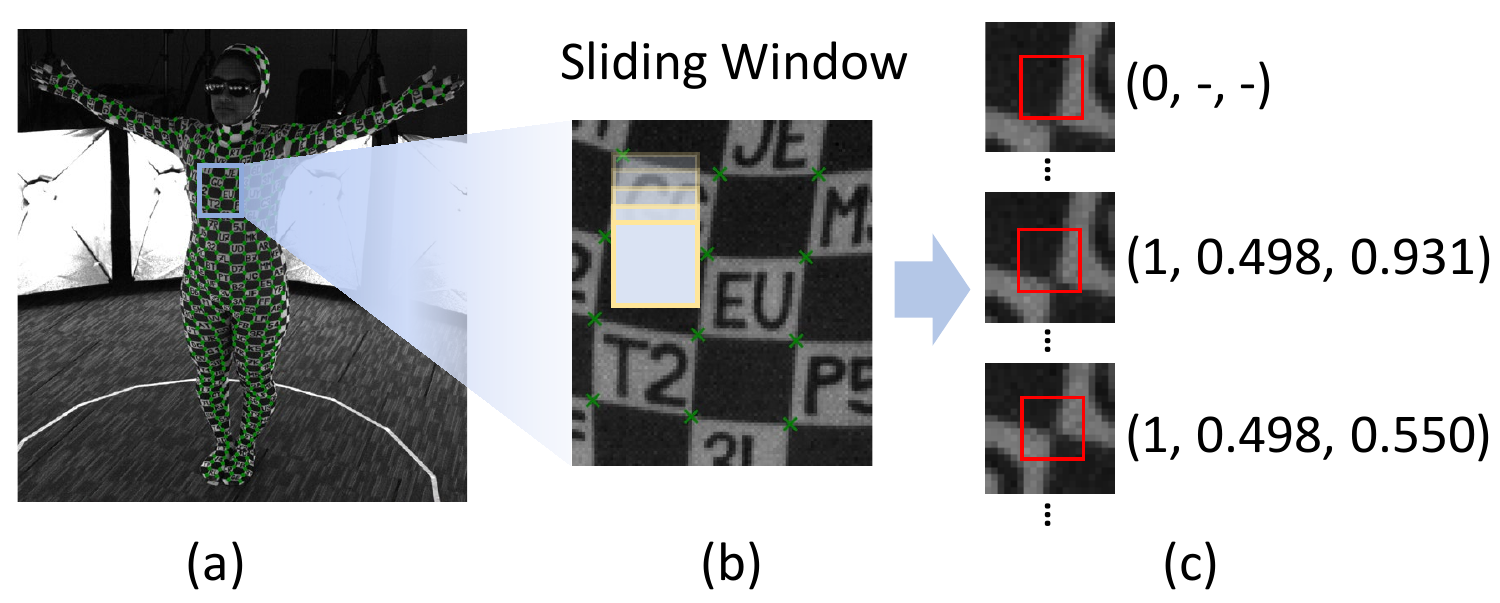}
    \caption{Generating training data for \Cornerdet{}: (a) The annotated image. (b) Sliding a $20\times20$ window across the annotated image. For each position of the window we generate a crop -- input for  \Cornerdet{}. (c) The crop is a positive sample (1, $x$, $y$) for  \Cornerdet{} if it contains a valid checkerboard corner in its center $8\times8$ pixels (red square); otherwise it is a negative sample (0, -, -). The first number is a binary value representing whether the crop is a positive sample, and the last two numbers are the normalized corners coordinates relative to the center $8\times8$ pixels.}
    \label{fig:CornerdetNetTrainingData}
\end{figure}

\paragraph{Corner Detector}
We generate the training data for \Cornerdet{} by sliding a $20\times20$ window with stride 1, as shown in Fig. \ref{fig:CornerdetNetTrainingData}b. \Rv{Each position of the $20\times20$ window will be cropped as} an input to \Cornerdet{}, labeled positive if and only if an annotated corner lies inside its center $8\times8$ pixels. For positive samples we also compute the sub-pixel corner coordinates relative to the center \Rv{$8\times8$ pixels}.

\paragraph{Quad Classifiers}
\begin{figure}
    \centering
    \includegraphics[width=\columnwidth]{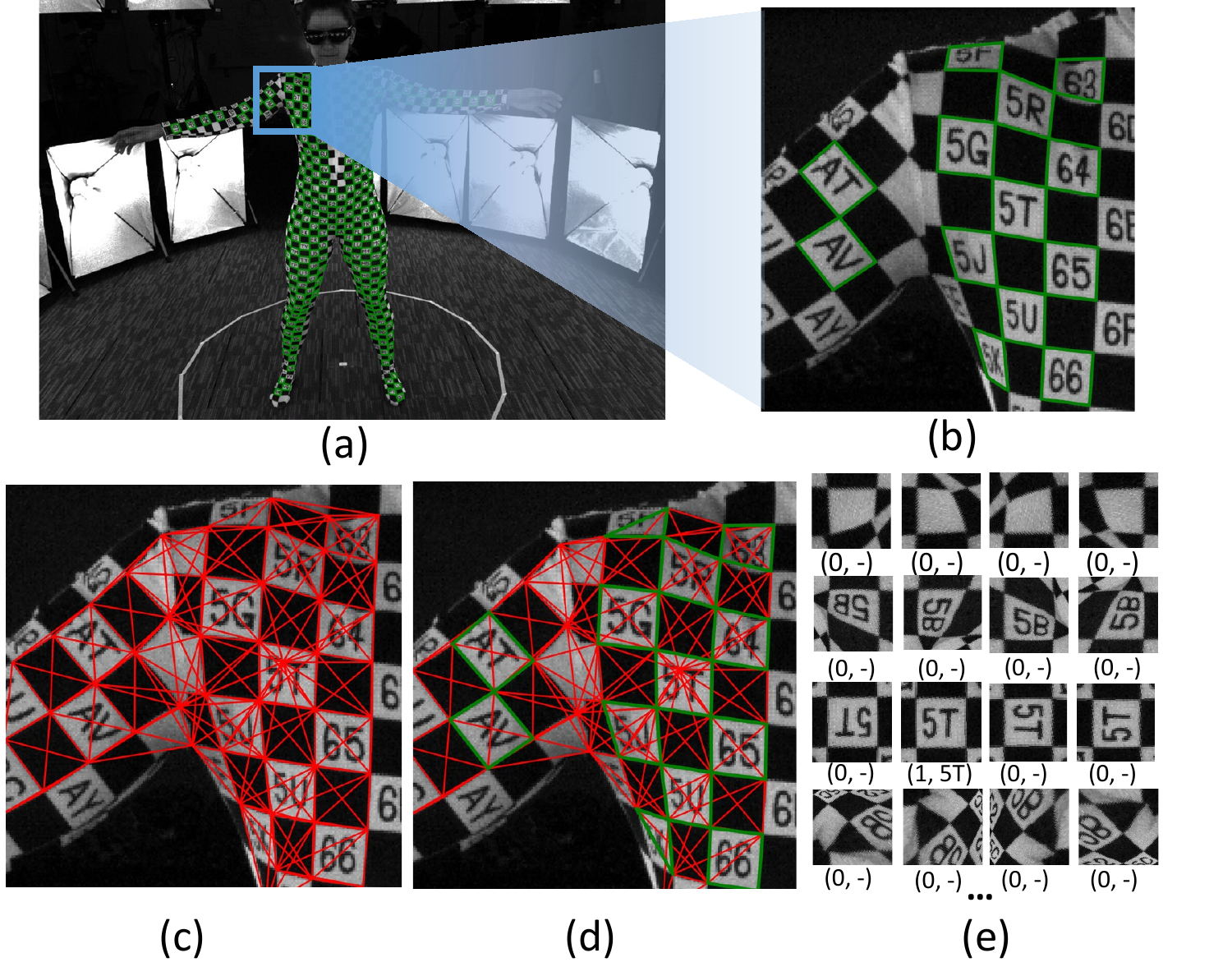}
    \caption{(a, b): Manual quad annotations. (c) The candidate quads generated using the algorithm from Section~3.2. (d) Selection of valid quads. (e) Homography-transformed quads with four possible rotations, including ground truth labels for training \Rejector{} and \RecogNet{}.
    }
    \label{fig:QuadClassifiersData}
\end{figure}
We start by generating candidate quads \ACChange{from the annotated corners in the manually annotated images } using the algorithm discussed in Appendix \ref{appendix:QuadClassifiers}. Note that the same quad generation algorithm will be used during deployment, i.e., when processing new motion sequences. The quad generator is conservative and creates many quads that do not correspond to valid white squares, see \Fig{fig:QuadClassifiersData}c. However, we know which quads are valid because all of the valid ones were manually annotated, see \Fig{fig:QuadClassifiersData}a. This allows us to automatically generate both positive and negative examples for a given candidate-quad generator, see \Fig{fig:QuadClassifiersData}d. These $104 \times 104$ images are used to train \Rejector{}. It is important \ACChange{that} the quad generator used during deployment is identical to the quad generator used when generating the training data for \Rejector{}. The two-letter code annotations of the valid quads are then used to train \RecogNet{}.

\subsection{{Data Augmentation and Synthetic Data}}
\label{appendix:DataAugmentationAndSyntheticData}
\begin{figure}
    \centering
    \includegraphics[width=\columnwidth]{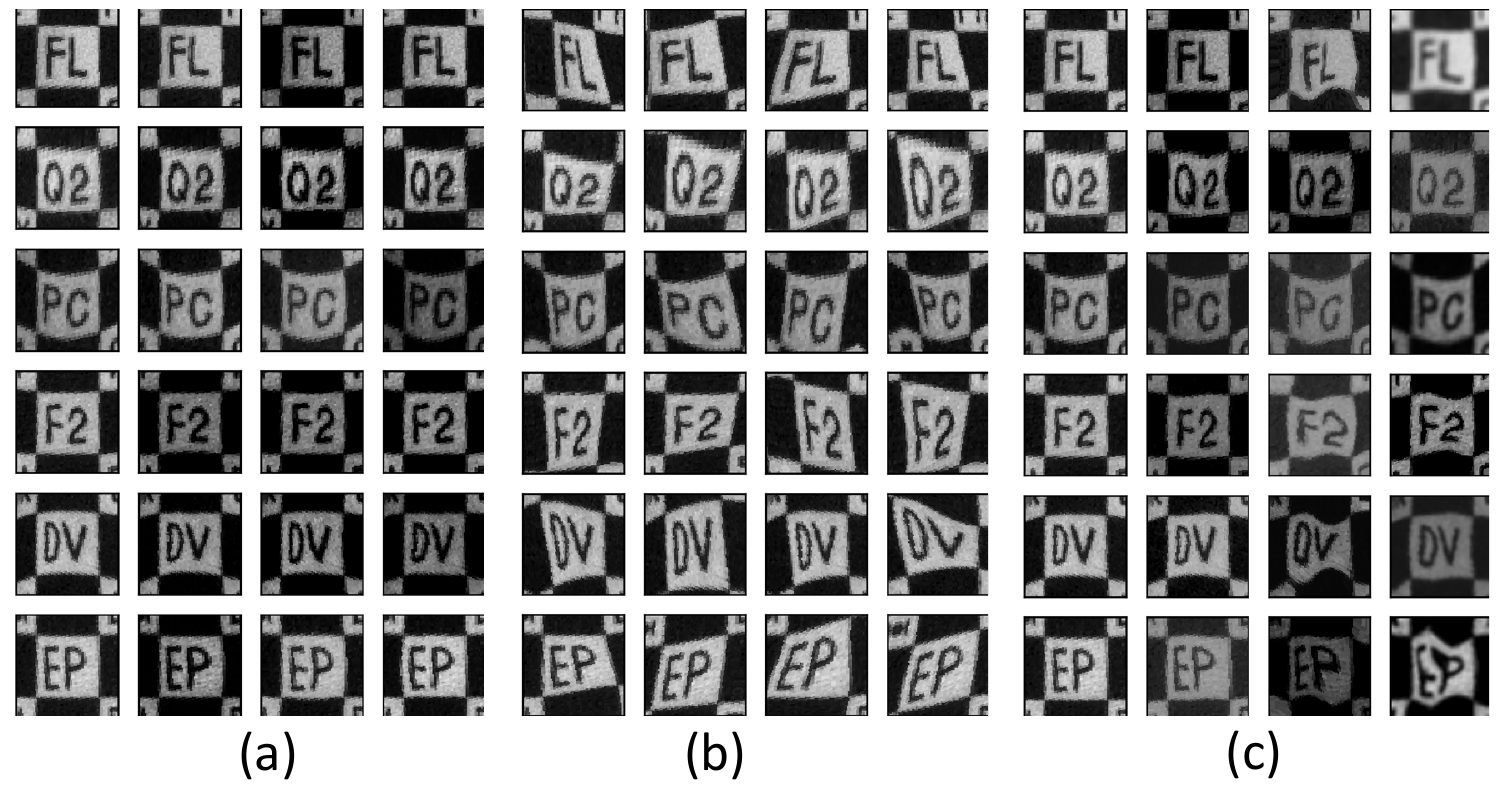}
    \caption{(a) The augmented \ACChange{positive} training data for \Rejector{}. (b) We generate negative training samples for \Rejector{} by warping one or more corners of a positive sample away from its original location, which simulates the case that the quad's corners were not correctly localized. (c) The data augmentation for \RecogNet{} is aggressive, including significant blurring and large elastic deformations.
    }
    \label{fig:QuadClassifiersDataAug}
\end{figure}
\label{subsec:DataAug}
\paragraph{Data Augmentation} 
All of the crops generated from annotated images as described in the Appendix \ref{appendix:DataConversion} are augmented by applying intensity perturbations (contrast, brightness, gamma). In addition, we also apply geometric deformations on each input \ACChange{image}. For the corner detector, we also augment the training data by generating random rotations of each \ACChange{image}, because checkerboard-like corners are rotation invariant.

Different data augmentation approaches need to be applied to \Rejector{} and \RecogNet{}. For the \Rejector{}, we blur the image using Gaussian filter and add elastic deformation using thin-plate splines \cite{wood2003thin} to simulate skin deformation. We constrain the elastic deformations to fix the checkerboard-like corners in place, see Fig. \ref{fig:QuadClassifiersDataAug}a, otherwise positive examples could be turned into negative ones. We also use this fact to our advantage: if we displace a checkerboard-like corner of a valid white square, we obtain a new (augmented) negative example, simulating the case when quad's corners have not been correctly localized, see Fig. \ref{fig:QuadClassifiersDataAug}b.

Since \RecogNet{} is required to predict characters from any input image, we can afford to augment our data more aggressively. Specifically, we use much more significant geometric distortions, intensity variations, blurring and \ACChange{additional} noise, see Fig.~\ref{fig:QuadClassifiersDataAug}c. This aggressive data augmentation has an interesting effect: the performance on the training data becomes worse, since we made the recognition task more difficult. However, we obtain better performance on the \textit{test} set, which is what matters. This agrees with human intuition: if students are given harder homework (training), they will likely perform better in their first job.

\paragraph{Synthetic Data}
To further enhance the diversity of our training data, we also generated synthetic data sets by rendering an animated SMPL \cite{loper2015smpl} model. We use synthetic data only for training the \RecogNet{}, because this was the bottleneck in the overall pipeline, see Section~6 for more details.
We textured the body mesh with the same checkerboard-like pattern as used in the real suit and applied animations from a public motion capture database \cite{AMASS:ICCV:2019}. We randomly generated new two-letter codes, including variations in font types and sizes to emulate \ACChange{the} handwriting of the codes. For each animation frame, we rendered images with virtual cameras, simulating our real capture setup by copying the intrinsic and extrinsic parameters from our real cameras. 
The visibility of corners in the rendered images is determined using ray tracing. To control the quality of quads that will be added to the training set, we check for corner visibility and use a classifier considering the quad’s 3D normal direction and quad geometry in the rendered image. 

\end{document}